\documentclass[journal]{IEEEtran}

\usepackage{makecell} % for making the table
\usepackage{fontspec}
\usepackage{varwidth} 
\usepackage{lipsum}
\usepackage{multicol}
\usepackage{multirow}
\usepackage{booktabs}
\usepackage{mwe}
\usepackage{enumerate}
\usepackage{titlesec}
\usepackage{verbatim}
\usepackage{overpic}
\usepackage{indentfirst}
\usepackage[dvipsnames,HTML]{xcolor}
\usepackage{adjustbox}
\usepackage{color, colortbl}
\usepackage{caption}
\usepackage{subcaption}
\usepackage{booktabs}
\usepackage{tabularx}                   
\usepackage{threeparttable}
\usepackage{amssymb}
\usepackage{amsfonts}

\usepackage{float}
\usepackage{soul}
\usepackage{comment}
\usepackage{amsmath}
\usepackage{graphicx}

\begin{document}

\title{Efficient Motion Sickness Assessment: Recreation of On-Road Driving on a Compact Test Track}

\author{Hüseyin~Harmankaya, Adrian Brietzke, Rebecca Pham Xuan, Barys Shyrokau, Riender Happee and Georgios~Papaioannou*
\thanks{Hüseyin~Harmankaya, Barys Shyrokau, Riender Happee and Georgios Papaioannou are with the Cognitive Robotics Department, Delft University of Technology (E-mail:g.papaioannou@tudelft.nl).
Rebecca Pham Xuan and Adrian Brietzke are with the Volkswagen AG,  Group Innovation, Germany.}
}

\markboth{ARXIV PREPRINT 2024}%
{Shell \MakeLowercase{\textit{et al.}}: Bare Demo of IEEEtran.cls for IEEE Journals}

\maketitle

\begin{abstract}

The ability to engage in other activities during the ride is considered by consumers as one of the key reasons for the adoption of automated vehicles. 
However, engagement in non-driving activities will provoke occupants' motion sickness, deteriorating their overall comfort and thereby risking acceptance of automated driving.
Therefore, it is critical to extend our understanding of motion sickness and unravel the modulating factors that affect it through experiments with participants. 
Currently, most experiments are conducted on public roads (realistic but not reproducible) or test tracks (feasible with prototype automated vehicles).
This research study develops a method to design an optimal path and speed reference to accurately replicate on-road motion sickness exposure on a small test track.
The method uses model predictive control to replicate the longitudinal and lateral accelerations collected from on-road drives on a test track of 70 m by 175 m.
A within-subject experiment (47 participants) was conducted comparing the occupants' motion sickness occurrence in test-track and on-road conditions, with the conditions being cross-randomized. 
The results illustrate that the subjective (reported) motion sickness is well reproduced with an insignificant reduction on the track. 
Meanwhile, there is an overall correspondence of individual sickness levels between on-road and test-track.
This paves the path for the employment of our method for a simpler, safer and more replicable assessment of motion sickness.

\end{abstract}


\section{Introduction}

The current developments of automated vehicles (AVs) are expected to change how occupants use their commute time.
Questionnaires have shown that users are keen on performing non-driving related tasks (NDRTs) during their travel with AVs \cite{kyriakidis2015public,nordhoff2019impressions}.
However, 60-70$\%$ of the population is expected to be affected by motion sickness (MS) with the envisaged scenarios of AVs (limited view on the horizon, backwards seating, etc. \cite{diels2016self,schmidt2020international,papaioannou2025occupants}). 
Experiments have shown that MS is caused in particular by low frequency motions and the absence of external (out of the window) visual information. While the physiological mechanisms underlying the development of MS remain unclear, experiments and simulations have shown that MS can be well explained by the sensory conflict theory, which states that MS results from conflict between expected and perceived motion \cite{Reason1978a}. 
Extensive research is being performed to understand the occurrence of MS and develop mitigation strategies \cite{emond2024will,yunus2025review}.
Both require thorough human experiments representative of on-road driving to capture various scenarios and triggers for MS modeling and to test the effectiveness of vehicle control strategies in mitigating motion sickness.
However, on-road driving experiments on public roads are potentially unsafe and less replicable, e.g. due to traffic and other road users.
Large test tracks are a viable option \cite{jones2019motion,brietzke2021influence}; however, they are relatively expensive, and the complexity of on-road driving is missed.

In AVs, the motion planner layer is responsible for identifying the optimal trajectory and behaviour in relation to surrounding traffic and infrastructure.
Due to the importance of MS in AVs, significant research has been conducted to ensure that the motion planner can derive a comfortable driving experience that can secure occupants' engagement in NDRTs. 
To that end, motion planning algorithms have been developed to consider MS accumulation as an objective for determining a nausea-free trajectory. 
A motion planning strategy has been proposed \cite{htike2020motion,htike2021fundamentals} that minimises the Illness Rating \cite{ISO2631} for longitudinal and lateral accelerations without compromising the journey time.
The above strategy has been adopted and employed as method for sorting motion planner alternatives based on multi-criteria evaluation, also considering metrics about ride comfort, aggressive driving, riding confidence,  energy efficiency, and others \cite{papaioannou2022multi}. 
However, these works employed offline optimization software (\textit{GPOPs-II} \cite{patterson2014gpops}) without real-time testing. 
This strategy was later modified and implemented in real-time using ForcesPro \cite{FORCESNLP} and also tested with human-in-the-loop experiments, mitigating MS by even 65$\%$ \cite{jain2023optimal}.
To model MS more accurately within the control modules, a frequency-shaping-based motion planning algorithm is proposed  \cite{li2021mitigating}, by minimising the motion sickness dose value (MSDV) of the frequency-weighted longitudinal and lateral accelerations.
A further step in this direction is presented in \cite{zheng2023mitigation}, proposing a receding-horizon formulation for optimization-based motion planning and also using deep reinforcement learning to minimize low-frequency accelerations \cite{rajesh2023comfort}.
Despite the importance of human body motion on motion comfort \cite{papaioannou2023impact,papaioannou2025occupants,messiouekastos}, limited studies have considered it within the control modules.
In this direction, a trajectory planning algorithm has been proposed \cite{steinke2022trajectory} incorporating passive head tilting and the subjective vertical conflict model. 
Based on the knowledge from these studies, this work has the objective of recreating on-road sickness exposure on test tracks.

The challenge of recreating motion sickness exposure on a compact test track is similar to the challenge of realistic motion cueing in driving simulators. 
In driving simulators, so-called motion filters or washout filters are routinely applied to create simulator platform motions fitting within the static and dynamic range of the simulator \cite{stahl2014state}.
Such a motion cueing algorithm is used \cite{talsma2023validation} to test the replication of sickness exposure from an on-road drive to six degrees of freedom moving base driving simulator, where they found the MS to be more severe for the on-road scenario compared to the driving simulator.
This shows a gap in capturing motion sickness exposure, limiting the use of driving simulators as a full replacement for on-road MS experiments
\cite{dam2024verifying}.
Traditional motion cueing algorithms used in driving simulators are computationally efficient but often do not fully use the available workspace. 
A more effective solution is the use of model predictive control. 
MPC can use the workspace more effectively but it is computationally demanding, making it challenging but not impossible to use MPC in real-time \cite{chadha2023computationally}. 
In this study, we use offline MPC; therefore, computational efficiency is not prohibitive. 
Where driving simulators typically offer motion with six degrees of freedom, we control vehicle longitudinal and lateral motion.

This study presents a method to recreate any on-road driving scenario on a compact test track (i.e., small size) for efficient and reproducible human in the loop motion sickness assessment (i.e., simplified test procedure of motion sickness assessment). 
The effectiveness of the proposed method in replicating the occurrence of on-road motion sickness has been assessed via human experiments.
The participants performed a NDRT during the experiment, ensuring limited exposure to the different visual environments between on-road and test-track driving. 
Their focus on the NDRT was assessed using a performance quantification method.
This ensured participants’ exposure to an environment similar to that of an AV and eliminated any additional sickening effects of the sensory conflicts arising through visual cues.

The contributions of the paper are as follows:
\begin{itemize}
    \item A novel method is proposed to recreate any on-road sickness occurrence in a compact test track, simplifying and enhancing the efficiency of human in the loop motion comfort assessment. The method uses model predictive control (MPC) to replicate the longitudinal and lateral accelerations collected from on-road drives. 
    \item Extensive within-subject human participants experiments on public roads and replication on a test track, demonstrate the effectiveness and validity of our proposed method. 
    Results prove that only a small controlled area is needed to conduct motion sickness experiments by employing the proposed method. 
\end{itemize}

This paper is structured as follows. 
Section \ref{sec:sim_models} describes the motion planning algorithm; 
Section \ref{sec:experiment} describes the experimental design of the evaluation study; 
Section \ref{sec:results} presents the results;
Section \ref{sec:conclusion} illustrates the conclusions and future work.

%%%%%%%%%%%%%%%%%%%%%%%%%%%%%%%%%%%%%%%%%%

\section{Motion planner algorithm}

\label{sec:sim_models}
This section covers the different elements of the algorithm developed to map naturalistic road sickness exposure on a compact test track. 
Firstly, the vehicle dynamics model is given. 
Then, the MPC framework is presented. 
Finally, the objective MS metrics used to evaluate the different iterations of the algorithm are described.

\subsection{Vehicle model}
\label{sec:vehicle_model}

\begin{figure}
    \centering
    \includegraphics[width=0.75\linewidth]{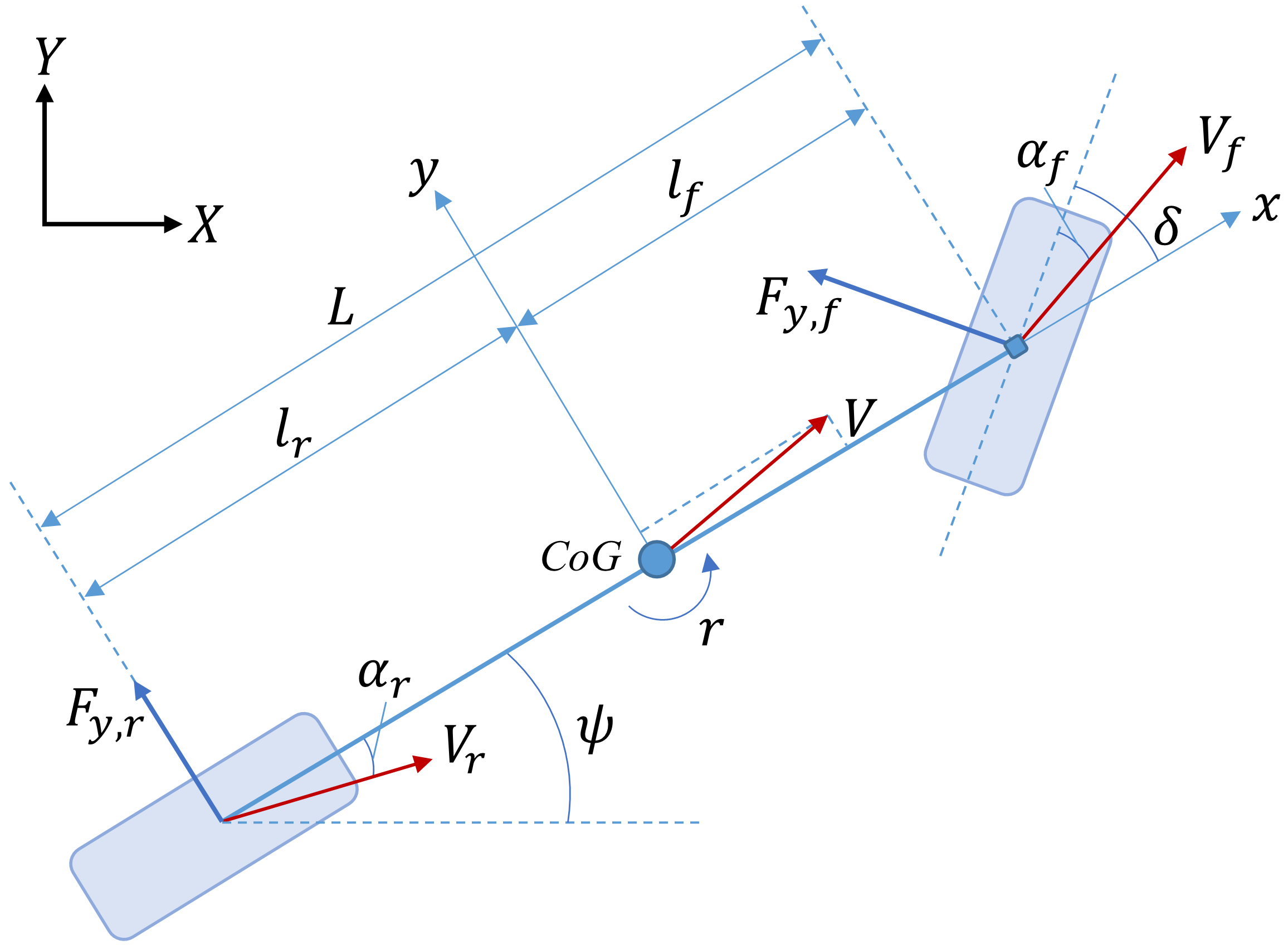}
    \caption{Bicycle model}
    \label{fig:bicycle_model}
\end{figure}

To efficiently capture the vehicle dynamics in the MPC-based motion planner, a linear bicycle model (Figure \ref{fig:bicycle_model}) is used: 

\begin{subequations}
    \begin{equation}
        \dot{X} = v_x \cos{(\psi)} - v_y \sin{(\psi)}
    \end{equation}
    \begin{equation}
        \dot{Y} = v_x \sin{(\psi)} + v_y \cos{(\psi)}
    \end{equation}
    \begin{equation}
        \dot{v}_x = a_x - \frac{F_{y,f}  \sin(\delta)}{m}  + v_y  r
    \end{equation}
    \begin{equation}
        \dot{v}_y = \frac{ F_{y,f} \cos{(\delta)} + F_{y,r} }{m} - v_x r
    \end{equation}
    \begin{equation}
        \dot{r} = \frac{ l_f F_{y,f} \cos{(\delta)} - l_r F_{y,r} }{I_{zz}}
    \end{equation}
\end{subequations}

\noindent where $X$ and $Y$ are the global coordinates; 
$v_x$ and $v_y$ are the longitudinal and lateral velocities; 
$a_x$ denotes the longitudinal acceleration;
$\psi$ is the heading angle;
$r$ is the yaw rate; 
$\delta$ the steering angle;
$m$ denotes the mass of the vehicle;
$I_{zz}$ is the moment of inertia;
$l_f$ is the distance between centre of gravity (CoG) and the front axle;
$l_r$ is the distance between CoG and the rear axle.

The lateral acceleration $a_y$ is determined under the assumption of steady-state turning and is described as \cite{matute2019experimental}:

\begin{equation}
    a_y = \frac{v_x^2 \delta}{l_f + l_r}.
\end{equation}

The lateral forces on the front and rear axles ($F_{y,f}$ and $F_{y,r}$) are determined with the assumption of linear tyre behaviour:

\begin{equation}
    F_{y,f} = C_{\alpha,f} \tan{(\alpha_f)}
\end{equation}

\begin{equation}
    F_{y,r} = C_{\alpha,r} \tan{(\alpha_r)}
\end{equation}

\noindent where $C_{\alpha,f}$ is the cornering stiffness of the front tyres and $C_{\alpha,r}$ of the rear tyres.
The slip angles at the front and rear axle ($\alpha_f$ and $\alpha_r$) are described as:

\begin{equation}
    \alpha_f = \delta - \frac{v_y + l_f r}{v_x}
\end{equation}

\begin{equation}
    \alpha_r = - \frac{v_y - l_r r}{v_x} .
\end{equation}

Thus, the state vector is expressed as:

\begin{equation}
    \textbf{x} = \begin{bmatrix} X & Y & v_x & v_y & \psi & r & \delta & a_x  \end{bmatrix}^T 
\end{equation}

\noindent while the control input vector is defined as:

\begin{equation}
    \textbf{u} = \begin{bmatrix} d\delta &  da_x \end{bmatrix}^T 
\end{equation}

\noindent where $d\delta$ is the steering rate, and $da_x$ is the longitudinal acceleration rate (jerk).

\subsection{Model Predictive Control}
\label{sec:mpc}
% This section covers the formulation of the MPC framework.
The MPC algorithm optimizes a cost function by predicting the vehicle's state over a defined horizon.
This allows the controller to find an optimal control input, while future states are considered.

\subsubsection{Problem formulation}

In this work, the objective of the MPC-based motion planning algorithm is to replicate the on-road motion sickness exposure in a compact test track by tracking the longitudinal and lateral accelerations from the on-road drive on the area of the compact track while considering various feasibility constraints.
Specifically, the algorithm, implemented in ForcesPro \cite{FORCESNLP}, considers the vehicle dynamics equations and the limits of the vehicle actuators, such as the maximum steering angle.
Additionally, the boundaries of the test track are taken as constraints, ensuring the vehicle stays within the designated area.

\subsubsection{Cost function}
\label{sec:cost_func}
The MPC controller is formulated as:

\begin{subequations}
\begin{align}
    \underset{\textbf{u}}{\min} \quad & J(\textbf{x},\textbf{u}) \\
    \text{s.t.:} \quad & \textbf{x}_{k+1}=f(\textbf{x}_k,\textbf{u}_k),\quad k=0,\dots,N_p-1 \\
    & \textbf{x}_{low,lim} \leq \textbf{x}_k \leq \textbf{x}_{up,lim},\quad k=0,\dots,N_p \label{eq:constr_x} \\
    & \textbf{u}_{low,lim} \leq \textbf{u}_k \leq \textbf{u}_{up,lim},\quad k=0,\dots,N_p \label{eq:constr_u} \\
    & \textbf{x}_0 = \textbf{x}_{init} .
\end{align}
\end{subequations}

The cost function $J$ is defined as:

\begin{equation}
\begin{split}
    J  = \sum_{k=0}^{N_p} & w_{a_x,k} (a_{x,k} - a_{x,ref,k})^2 \\
     + & w_{a_y,k} (a_{y,k} - a_{y,ref,k})^2 \\
     + & w_{X,k} (X_k - X_{centre})^2\\
     + & w_{Y,k} (Y_k - Y_{centre})^2 \\
     + & w_{d\delta} d\delta_k + w_{da_x} da_{x,k}
\end{split}
\end{equation}

\noindent where $N_p$ is the prediction horizon set at 90 timesteps with the sampling time of $T_s=0.1s$. 
These values ensure realistic motion over the frequencies relevant for motion sickness (0.1-1 Hz). 
$a_{x,ref}$ and $a_{y,ref}$ are the reference longitudinal and lateral accelerations recorded from the on-road drive;
$w_{a_x}$ and $w_{a_y}$ are the adaptive weights on the error for the longitudinal and lateral accelerations.
To ensure that the vehicle is steered back when getting close to the edges of the area, the offset from the centre of the track $(X_{centre},Y_{centre})$ is penalized with adaptive weights $w_X$ and $w_Y$ for the $X$ and $Y$ position errors, respectively.
Lastly, $w_{d\delta}$ and $w_{da_x}$ are the weights on the control inputs steering rate and longitudinal acceleration rate, respectively.

To secure the feasibility of the solution, inequality constraints,  as specified in Equations \ref{eq:constr_x} and \ref{eq:constr_u}, are introduced with the following boundaries:

\begin{equation}
    \begin{bmatrix}
        0 \text{m}\\
        0 \text{m}\\
        1 \frac{\text{m}}{\text{s}}\\
     %   -\infty \\ % -\text{inf}
     %   -\infty \\
     %   -\infty \\
        -20 \deg\\
        -4.1 \frac{\text{m}}{\text{s}^2}
    \end{bmatrix}
    \le
    \begin{bmatrix} 
    X \\
    Y \\ 
    v_x\\
    %v_y \\
    %\psi \\ 
    %r \\ 
    \delta \\
     a_x  \\
     \end{bmatrix} 
    \le
    \begin{bmatrix}
        175 \text{m}\\
        70 \text{m}\\
        11.1 \frac{\text{m}}{\text{s}}\\
     %   \infty\\
     %   \infty\\
     %   \infty\\
        20 \deg\\
        2.5 \frac{\text{m}}{\text{s}^2}
    \end{bmatrix} 
\end{equation}

\begin{equation}
    \begin{bmatrix}
        -14.4 \frac{\deg}{\text{s}} \\
        -4.1 \frac{\text{m}}{\text{s}^3}
    \end{bmatrix}
    \le
    \begin{bmatrix}
    d\delta \\
    da_x 
    \end{bmatrix}
    \le
    \begin{bmatrix}
        14.4 \frac{\deg}{\text{s}} \\
        2.3 \frac{\text{m}}{\text{s}^3}
    \end{bmatrix}.
\end{equation}

\noindent The upper and lower boundaries of the inequality constraints have been selected based on the dimension of the test area and the vehicle's limits. 
The speed limits ($v_x$) were defined based on requirements from the test-track operator.
The initial states are defined as follows:

\begin{equation}
\begin{split}
    \textbf{x}_{init} & = 
    %\begin{bmatrix} X_0 & Y_0 & v_{x,0} & v_{y,0} & \psi_0 & r_0 & \delta_0 & a_{x,0}  \end{bmatrix}^T \\
     \begin{bmatrix} 15 \text{m} & 65 \text{m} & 2 \frac{\text{m}}{\text{s}} & 0 & 0 & 0 & 0 & 0  \end{bmatrix}^T 
\end{split}
\end{equation}

\noindent The vehicle initiated at the top left corner of the test track (Figure \ref{fig:paths}) to provide the maximum available space to the motion planner. 

\subsubsection{Adaptive weights}

To fully utilize the test area and avoid the vehicle driving beyond the test track edges, the acceleration weights $w_{a_x}$ and $w_{a_y}$ consider adaptive coefficients which depend on the vehicle offset from the centre of the area.
As a result, the weights become lower when the vehicle is closer to the edges of the test area, decreasing the priority of accurately tracking the longitudinal and lateral accelerations. 
The adaptive weights are defined as:

\begin{equation}
    i_{norm,k} = \frac{i_k - i_{centre}}{i_{centre}}, i = \{ X, Y \}
\end{equation}

\begin{equation}
    c_{j,k} = 1 - 0.99 i_{norm,k}^8, j = \{ x, y \}
\end{equation}

\begin{equation}
    w_{a_j,k} = c_{x,k} c_{y,k} w_{c,a_j}
\end{equation}

\noindent where $w_{c,a_x}$ and $w_{c,a_y}$ are the tuned constant weights set for the longitudinal and lateral accelerations.

Furthermore, to ensure the avoidance of going beyond the test track's boundaries and steering the vehicle back to the centre, the position weights $w_X$ and $w_Y$ also dynamically change depending on the offset from the centre of the test area.
The weights increase non-linearly as the vehicle gets closer to the boundaries of the track:

\begin{equation}
    w_{i,k} = \frac{w_{c,i}}{1 - i_{norm,k}^4} - w_{c,i}
\end{equation}

\noindent where $w_{c,X}$ and $w_{c,Y}$ are tuned constant weights for the global $X$ and $Y$ position offset from the centre of the area. 
The values determined for the weights are listed in Table \ref{tab:weights}.

\begin{table}[h]
    \centering
    \caption{Weight settings}
    \begin{tabular}{|c c||c c|}
        \hline
        Weight & Value & Weight & Value \\ \hline
        $w_{c,a_x}$ & 300 & $w_{c,a_y}$ & 500 \\
        $w_{c,X}$ & 0.05 & $w_{c,Y}$ & 0.25 \\ 
        $w_{d\delta}$ & 0.2 & $w_{da_x}$ & 0.2 \\ \hline
    \end{tabular}
    \label{tab:weights}
\end{table}

\subsection{Standstill sections}

The path-following controller implemented in the test vehicle is constrained to maintain the vehicle speed above 1$\frac{\text{m}}{\text{s}}$ while driving. 
As a result, standstills cannot be fully represented. 
To overcome this, standstills were implemented manually by determining the locations where they should happen to stop the vehicle at these locations. 
The stops were identified from the reference path, and the most occurring stops were then replicated.
Through the manual implementation of the stops, the effects of deceleration and acceleration for the standstill events during the on-road drive were better represented.

To accommodate these stops within the boundaries of the test area, the path of the predicted horizon is used as the path to perform the deceleration and acceleration when the stop has been identified.
For the implementation, firstly, the time to perform the acceleration or the deceleration is determined based on a desired $a_x$ (Eq. \ref{eq:t_d_a}), which is identified from the stops in the on-road data.
This value is rounded off to one decimal place to match the time steps of the MPC controller, allowing the acceleration or deceleration to fit properly within the timesteps.
Then, the time duration for the acceleration/deceleration is used to determine the exact value of $a_x$ to accelerate/decelerate from the initial to the desired speed:

\begin{equation}
    t_{decel/accel} = \text{round} \left( \frac{v_{x,end} - v_{x,start}}{a_{x,desired}} , 1 \right) 
    \label{eq:t_d_a}
\end{equation}

\begin{equation}
    a_{x,const} = \frac{v_{x,end} - v_{x,start}}{t_{decel/accel}}
\end{equation}

\noindent where $v_{x,start}$ is the velocity at which the acceleration/deceleration motion is initiated;
$v_{x,end}$ is the desired speed at the end of the motion.
The speed $v$ and displacement $s$ of the motion are determined with:

\begin{equation}
    a(t+1) = a(t) = a_{x,const}
\end{equation}

\begin{equation}
    v(t+1) = v(t) + a(t+1) T
\end{equation}

\begin{equation}
    s(t+1) = s(t) + v(t+1) T + \frac{1}{2} a(t+1) T^2
\end{equation}

\noindent where the displacement is then used to interpolate the new $X$ and $Y$ coordinates in the prediction horizon.

\section{Objective assessment of motion sickness}
\label{sec:ms_metrics}

ISO-2631:1997 \cite{ISO2631} provides guidelines for objectively measuring and evaluating human exposure to whole-body mechanical vibration and repeated shock. 
The guidelines suggest the consideration of two metrics representing: (1) ride comfort emphasizing the higher frequencies (mainly above 1 Hz); (2) MS emphasizing the lower frequencies (mainly below 1~Hz). 
The latter metric is the primary focus of this study. 
For the derivation of the MS metric, frequency weightings are applied to the six degrees of freedom motion, i.e., both three-dimensional translation and rotation.
The metrics are evaluated by combining the root mean square (RMS) values of weighted accelerations ($MS_{W_{i}}$), translational and rotational, measured at the vehicle's centre of gravity.
More specifically, the RMS value of each acceleration is calculated as follows:

\begin{equation}
\label{eq:accel}
	MSDV_{i}=  \bigg(\int_{0}^{t} a_{i_W}^2 d\tau \bigg)^{\frac{1}{2}}
\end{equation}

\noindent where $i$ is the acceleration type, either translational or rotational, while $a_{W_i}$ stands for the weighted accelerations in the time domain.
In this work, only the lateral ($i=y$) and longitudinal ($i=x$) accelerations are considered.
After multiplying each of the $MSDV_{i}$ by appropriate factors ($k_i$), they are summed and the overall MSDV$_{total}$ is calculated as:

\begin{equation}
\label{eq:RC}
MSDV_{total} = \bigg( \sum k_i^2 MSDV_{i}^2 \bigg)^{\frac{1}{2}} 
\end{equation}

\noindent where $k_i$ is a multiplying factor for each term. The values are used according to the guidelines in the literature \cite{ISO2631}.
For motion sickness, the weighting filters for longitudinal $WP_{f_x}$, lateral $WP_{f_y}$ and rotational motion $WP_{f_r}$ are used based on the literature \cite{papaioannou2023impact}.  
In this work, we initially calculated $MSDV_{total}$  by using all the translational and rotational accelerations, but only the longitudinal and lateral accelerations were eventually used due to their impact on motion sickness \cite{papaioannou2023impact}.

%%%%%%%%%%%%%%%%%%%%%%%%%%%%%%%%%%%%%%%%%%

\section{Experiment design} \label{sec:experiment}

\subsection{Overview}

To evaluate our algorithm's ability to replicate the on-road sickness exposure, a within-subject experiment (i.e., all participants experienced both conditions) was conducted. The condition order (on-road and test track) was cross-over randomized to be balanced equally across the participants.

In the experiment, the participants were seated on the front passenger seat and an experimenter was seated at the back to record the participants’ feedback.
Participants were tasked to rate their motion sickness symptoms verbally in both conditions.

\subsection{Ethics statement}

The experiment was performed in accordance with the Declaration of Helsinki.
The study was approved by the Human Research Ethics Council of Delft University of Technology (Delft, The Netherlands;
application number 3598). 
All participants gave their written informed consent prior to participation in the study. 
Participants received a compensation of 50 euros.

\subsection{Participants}

\begin{figure}
    \centering
    \includegraphics[width=0.9\linewidth]{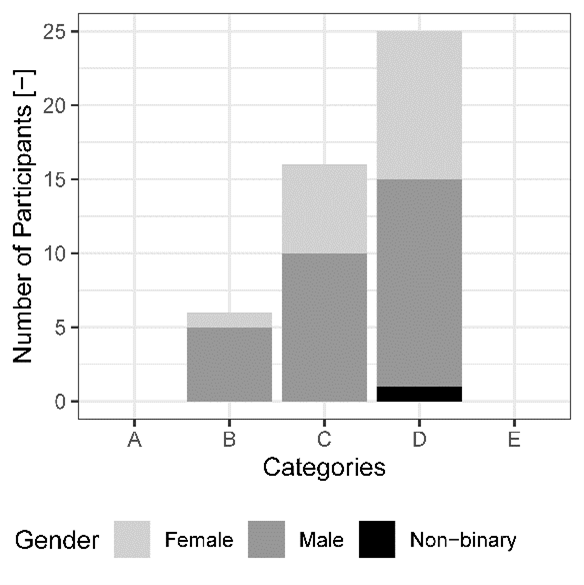}
    \caption{Categorization of the recruited participants in terms of gender and motion sickness susceptibility (A=lowest, E=highest susceptibility).}
    \label{fig:MSSQ}
\end{figure}

The study participants were recruited based on age, gender, and motion sickness susceptibility.
A wide range of ages (adults) was recruited. 
Motion sickness susceptibility was screened to avoid highly susceptible or insusceptible participants, and gender was used to create a gender-balanced dataset as much possible. 
The final sample consisted of 47 participants (29 males, 17 females, 1 non-binary). 
All the participants were healthy and without any vestibular dysfunction.
The age ranged between 17 and 68 (M = 27, SD = 12.17). % out of this sample 23 participants wore the XSENS motion capture body suit to record their body motion. 
Susceptibility was assessed using the motion sickness susceptibility questionnaire (MSSQ-Short) and by taking into special account the answer of the item regarding the experience of sickness in cars in the past 10 years \cite{xuan2023evaluation}.
Based on the answers, participants were categorized, and the ones falling within categories C to D were recruited to participate, while participants falling within category B were also invited as participants. 
Participants with very high susceptibility (Category E) were excluded.
The overview of the recruited participants is presented in Figure \ref{fig:MSSQ}.

General motion sickness susceptibility assessed prior to the experiment yielded a mean MSSQ-Short of 12.97 and standard deviation of 5.99. 
The mean MSSQ-Short across the participants translates to 55.90$\%$ percentile, which indicates that our participants have a slightly above-average motion sickness susceptibility.
The standard deviation translates to 44.96$\%$ to 65.57$\%$ percentiles, indicating that our participants cover a limited range of susceptibility.

\subsection{Apparatus}

\begin{figure}
    \centering
    \includegraphics[width=0.9\linewidth]{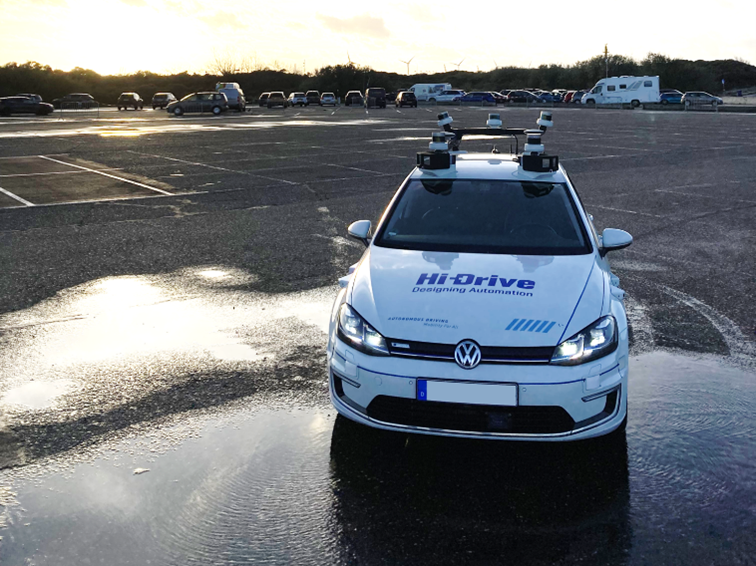}
    \caption{The test vehicle, a Volkswagen E-Golf equipped with experimental automation.}
    \label{fig:test_vehicle}
\end{figure}

As a test vehicle, a Volkswagen E-Golf equipped with an experimental SAE level 3 automated driving function was used (Figure \ref{fig:test_vehicle}). 
The vehicle was equipped with vehicle data collection devices with a 50-Hz sampling rate. 

\subsection{Procedure}

The on-road condition took place at Delft, the Netherlands and the test-track condition took place at a compact parking lot at Hoek van Holland, the Netherlands. 
During the on-road condition, the vehicle was manually driven by one of the three safety drivers following specific driving instructions to secure similarity between the on-road tests.
The route was a predetermined route on urban and interurban roads in medium to high traffic density.
The rides per safety driver were equally balanced ($\sim$15 per driver) in the on-road condition. 
The safety drivers were trained to maintain a predetermined maximum speed for various sections to minimize differences in driving styles.

The test vehicle was driven autonomously in the test-track condition, while the safety driver was monitoring the vehicle behind the steering wheel.
The driving style of the test-track condition was predefined using MPC. 
Hence, the applied motion was virtually identical for all participants.
The ride was around 25 minutes in both cases, which was sufficient for the accumulation of motion sickness.
The participants were invited in the second session with a gap of more than five days after the first trial to avoid habituation effects.
However, due to time limitations, a few participants were invited with a smaller gap between the two sessions ($\sim$ 1-2 days). 
The average gap across all participants was four days. 

During both conditions, the participants were instructed to focus on a non-driving related task, which was video watching.
To secure participants' constant engagement with the NDRT, videos were edited by including unexpected events about every 90 s. 
The participants were asked to count these events to eventually assess their engagement with the NDRT (NDRT\_counts).
The NDRT ensures participants’ exposure to an environment similar to that of an AV, where they expect to engage in NDRTs. 
Meanwhile, this eliminates any additional sickening effects of the sensory conflicts arising through uncontrolled out of the window visual cues.
The total number of NDRT\_counts was 20 and corresponded to the maximum counts in the video that could be identified during a complete ride (i.e., $\sim$25 min).

Regarding the motion sickness assessment, participants were instructed to report their sickness on the 11-point Motion Illness Symptoms Classification scale (MISC) \cite{reuten2021feelings}.
During the drives, every 1 minute, a beep was played within the car to motivate the participant to verbally state their MISC level \cite{reuten2021feelings}.
Additionally, participants were urged to self-report any changes on the scale in between minutes if the MISC rating changed since the last response was requested. 
Responses were recorded through the CAN system by the experimenter's input.
The drive was aborted if a MISC level of 6 was reached or if the participant requested to stop the experiment.
If no measurement has been recorded in a minute, the last MISC rating before that recorded was used. 
MISC ratings are filled at the level of the last rating per trial for the remaining measurement increments until the end of the ride.

%%%%%%%%%%%%%%%%%%%%%%%%%%%%%%%%%%%%%%%%%%

\section{Results: Part A - Motion Planning}
\label{sec:results}

This section covers the results of the motion planner algorithm, illustrating the generated path, its tracking performance and the objective comparison of motion comfort in four cases. 
The four cases considered are the following:

\begin{itemize}
    \item \textbf{Reference data}: The average path was calculated from five on-road drives performed by three safety drivers. 
    The velocity profile resulted by averaging the longitudinal and lateral accelerations. 
    %as illustrated in Figure \ref{fig:avrg_velocity}.
    The data collected from the vehicle for the on-road reference data are filtered using a zero-phase low-pass filter (Table \ref{tab:filter}), which maintains the frequencies relevant to MS and removes the noise.
    \item \textbf{On-road data}: This data is collected on the predetermined route. The same filtering method is applied as before.
    \item \textbf{Generated data (motion planner)}: The path is generated by the MPC framework, described in Section \ref{sec:sim_models}, by tracking the longitudinal and lateral accelerations from the on-road reference data, while adhering to the constraints of the test track. 
    The motion planner generates the trajectory with the velocity profile assigned to it.
    \item \textbf{Test track data}: The data is collected by the experimental vehicle when following the generated path from the motion planner. 
    The experimental vehicle uses its onboard path-tracking controller to follow the motion planner's generated path autonomously.
    The onboard controller was not modified, but used as provided by the project partner.
    The same filtering process as for the on-road reference data was applied in the test track data (Table \ref{tab:filter}).
\end{itemize}

\begin{table}[h!]
    \centering
    \caption{Zero-phase low-pass filter settings.}
    \begin{tabular}{|c|c|}
        \hline
        Setting & Value \\ \hline
        Sampling rate & 10Hz \\      
        Passband edge frequency & 0.05Hz \\
        Stopband edge frequency & 2Hz \\
        Passband ripple & 0.01dB \\
        Stopband attenuation & 160dB \\ \hline
    \end{tabular}
    \label{tab:filter}
\end{table}

The tracking performance is assessed through the comparison of the longitudinal and lateral accelerations of the three cases associated with the MPC approach (Reference data, Generated data and Test-track data)  in time (Figure \ref{fig:time_domain_acc_all}) and frequency domain (Figure \ref{fig:freq_plots1}).
The results focus on data collected from one drive per case.

\subsection{Generated path}

\begin{figure}[h]
\centering
\begin{subfigure}{1\linewidth}
  \centering
  \includegraphics[width=.90\linewidth]{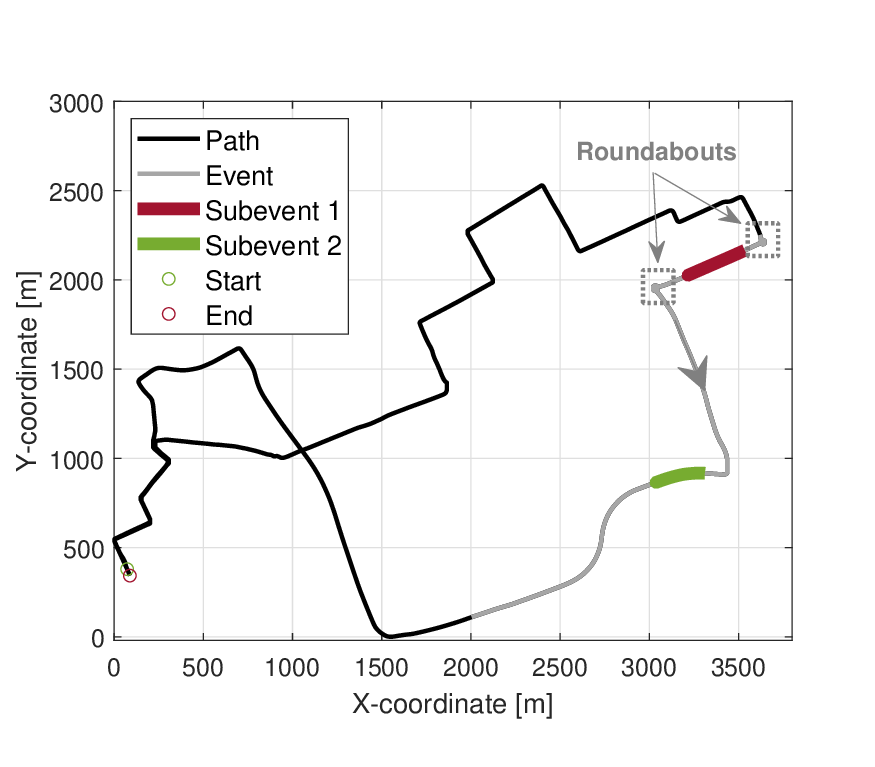}
  \caption{On-road path.}
  \label{fig:path_onroad}
\end{subfigure}%

\begin{subfigure}{1\linewidth}
  \centering
  \includegraphics[width=.90\linewidth]{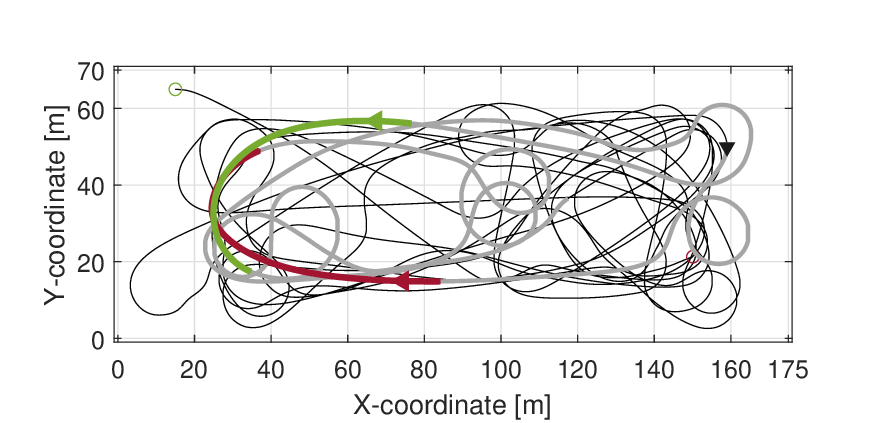}
  \caption{Test track path.}
  \label{fig:path_gen_track}
\end{subfigure}
\caption{Recorded paths for both on-road and test track scenarios.}
\label{fig:paths}
\end{figure}

The longitudinal and lateral accelerations from the on-road data are used as a reference signal to the MPC framework to generate a path to fit into the compact test track.
Figure \ref{fig:path_onroad} presents the on-road path and Figure \ref{fig:path_gen_track} illustrates the path measured on the test track, illustrating the difference in the total area covered for the two scenarios.
Figure \ref{fig:path_gen_track} also shows the algorithm avoiding the track boundaries by steering.

\subsection{Tracking performance}

To assess the replicability of the road sickness exposure, this section evaluates and compares the accelerations for each of the three cases: on-road, generated, and measured test track data.

\subsubsection{Time-domain analysis}

\begin{figure}[hp!]
\centering
\begin{subfigure}{\linewidth}
  \centering
  \includegraphics[width=\linewidth]{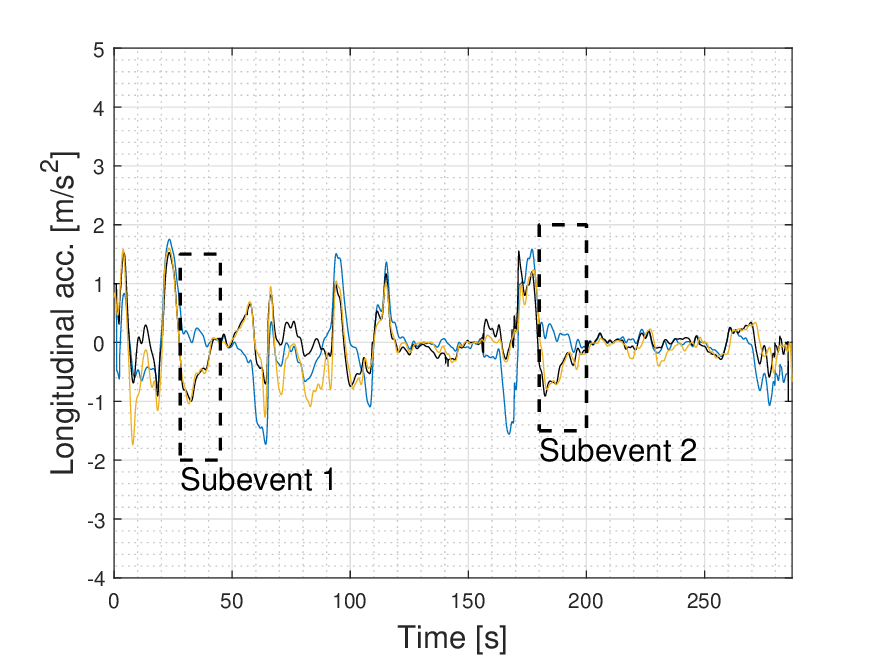}
  \caption{$a_x$.}
  \label{fig:ax_shifted02}
\end{subfigure}%

\begin{subfigure}{\linewidth}
  \centering
  \includegraphics[width=\linewidth]{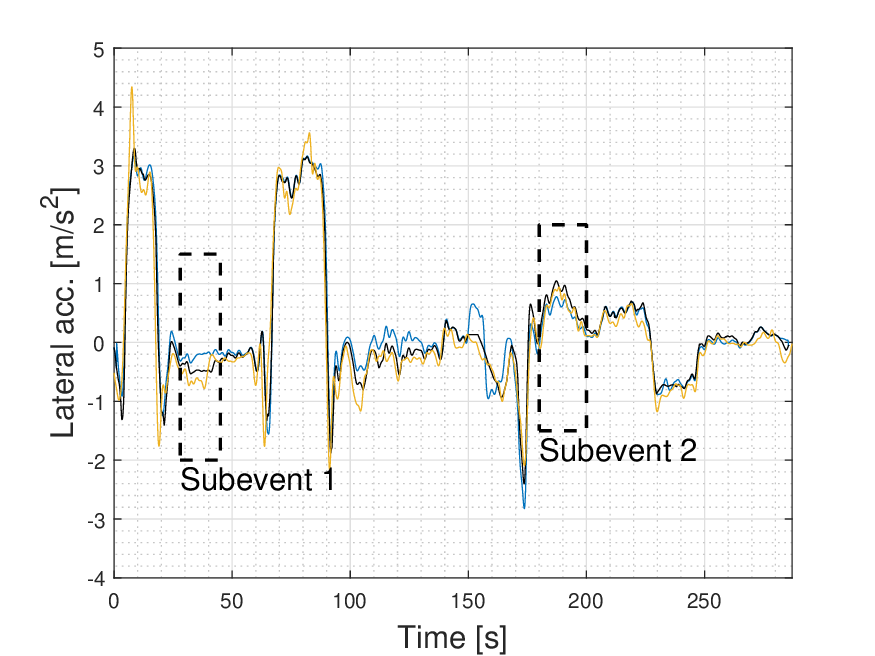}
  \caption{$a_y$.}
  \label{fig:ay_shifted02}
\end{subfigure}

\begin{subfigure}{\linewidth}
  \centering
  \includegraphics[width=\linewidth]{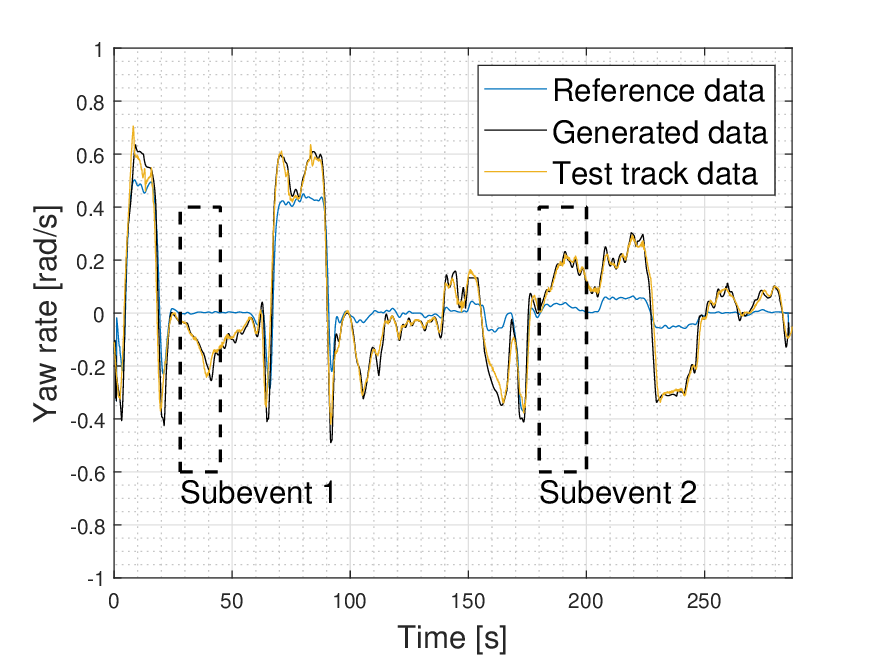}
  \caption{$r$ for one event.}
  \label{fig:r_shifted02}
\end{subfigure}
\caption{Comparison of event-based synchronized time domain longitudinal and lateral accelerations, and yaw rate between on-road reference motion, generated desired track motion, and measured test track motion. The test-track and road data correspond to Participant 33.}
\label{fig:time_domain_acc_all}
\end{figure}

The event-based time domain accelerations are presented in Figure \ref{fig:time_domain_acc_all} for the event that corresponds to the grey section illustrated in Figure \ref{fig:path_onroad}-\ref{fig:path_gen_track}. 
This includes two roundabouts, corners of different curvatures, and braking and accelerating events.

During the experiment in the test track, the experimental vehicle's path-tracking controller extended the standstill phases of the reference data for a longer period than the generated. 
This led to accumulating minor delays ($<$ 1 s) throughout the path for the recorded test track data.
To account for these delays, an event-based comparison of the time domain accelerations is conducted, where the longitudinal and lateral accelerations, and the yaw rate are synchronized based on the standstills (Figures \ref{fig:ax_shifted02}-\ref{fig:r_shifted02}).

The measured motion on the track very well matched the desired motion generated by the MPC illustrating good path tracking performance by the experimental vehicle.
Comparing the measured motion on the track and the on-road experiment, an almost perfect match was obtained for lateral acceleration, with a reasonable match for longitudinal acceleration. 
Yaw rate was well matched in the 2 corners around $t=$35s and $t=$80s whereas a somewhat higher yaw rate was needed on the track in later events. 
However, the average speed was much lower on the track (2.92 $\pm$ 0.08 m/s) compared to the on-road (8.79 $\pm$ 0.83 m/s), illustrating the ability of our motion planner to provide similar accelerations at much lower speeds.

According to Figure \ref{fig:ax_shifted02}, the motion planner's generated longitudinal acceleration deviates from the reference at two instances, i.e.,  at $t \approx 35s$ and $t \approx 185s$ (Subevent 1 and 2, Figure \ref{fig:path_gen_track} and \ref{fig:time_domain_acc_all}).
At these moments, the vehicle is close to the test track boundaries (Figure \ref{fig:path_gen_track}), while the reference lateral acceleration is relatively low (Subevent 1) or almost zero (Subevent 2) (Figure \ref{fig:ay_shifted02}).
As a result, the MPC determines that decelerating the vehicle, thus decreasing the velocity, is the optimal solution to achieve the desired lateral acceleration while also taking a turn to avoid the test track's boundaries.
As a result, the yaw rate (Figure \ref{fig:r_shifted02}) is significantly increased compared to the on-road data, where it was almost zero. 
This compromise occurs because the penalty by the adaptive weights on the offset from the centre of the track is increased urging the vehicle to return to the centre of the test track, while the priority on tracking the reference accelerations decreases.
Additionally, the balance of the weights on the accelerations is prioritizing the lateral acceleration more than the longitudinal (Table \ref{tab:weights}), hence the lateral acceleration is tracked better than the longitudinal.
This slightly higher priority on the lateral acceleration is due to the small area of the test track not allowing high speeds to be reached, thus not being able to fully track the on-road data.

Another interesting remark concerns the periods after the second roundabout and before Subevent 2 ($\sim$100-150$s$) and after Subevent 2 ($\sim$200-280$s$). 
During these periods, both the longitudinal and lateral accelerations are well tracked by the motion planner. 
However, due to the test track space limitations and the adaptive weights on the offset from the centre of the track, the motion planner urges the vehicle in a few directional changes increasing the yaw rate (Figure \ref{fig:r_shifted02}). 
So, on contrary with the tracking of Subevent 1 and 2, the directional changes in this time spans did not affect negatively the tracking of the reference signals, but increased the yaw rate. 

\subsubsection{Frequency-domain analysis}

%%%%%%%%%%%%%%%%%%%%%%%%%%%%%%%%%%%%%%%%%%%%%%%%%%%%%%%%%%%%%%

\begin{figure*}[hp!]
\centering
\begin{subfigure}{0.33\linewidth}
  \centering
  \includegraphics[width=\linewidth]{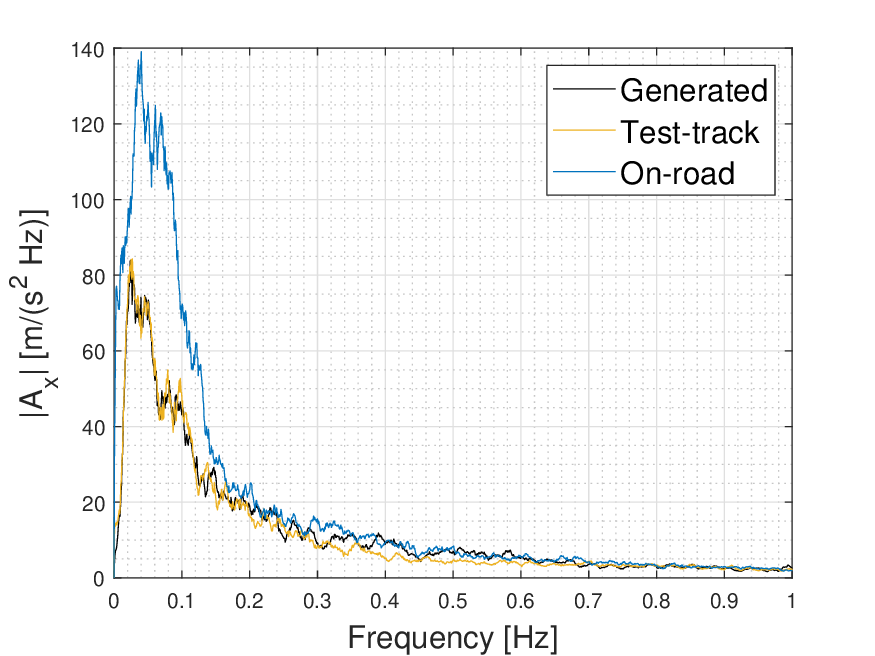}
  \caption{Unweighted $a_x$.}
  \label{fig:ax_unfilt}
\end{subfigure}%
\begin{subfigure}{0.33\linewidth}
  \centering
  \includegraphics[width=\linewidth]{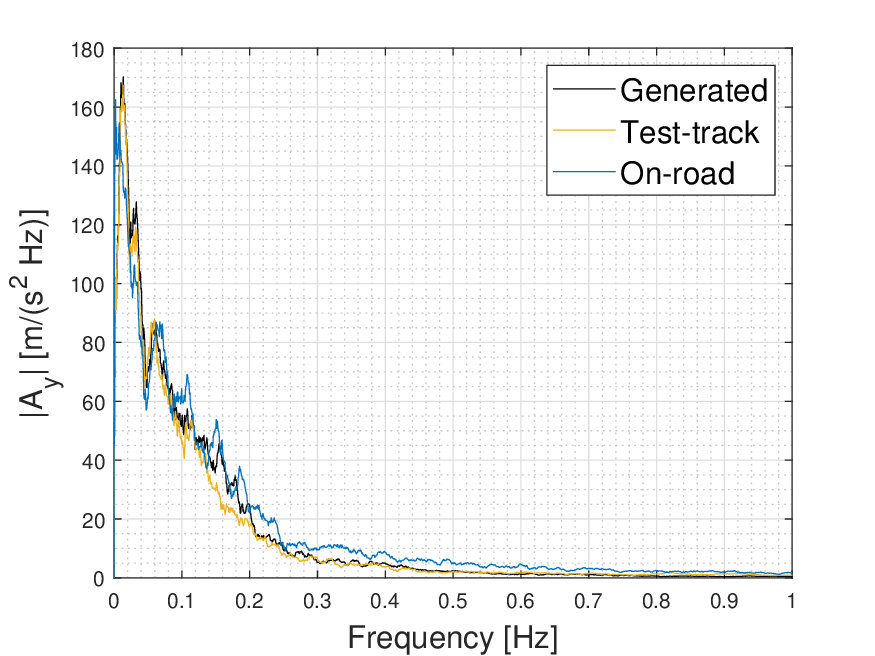}
  \caption{Unweighted $a_y$.}
  \label{fig:ay_unfilt}
\end{subfigure}%
\begin{subfigure}{0.33\linewidth}
  \centering
  \includegraphics[width=\linewidth]{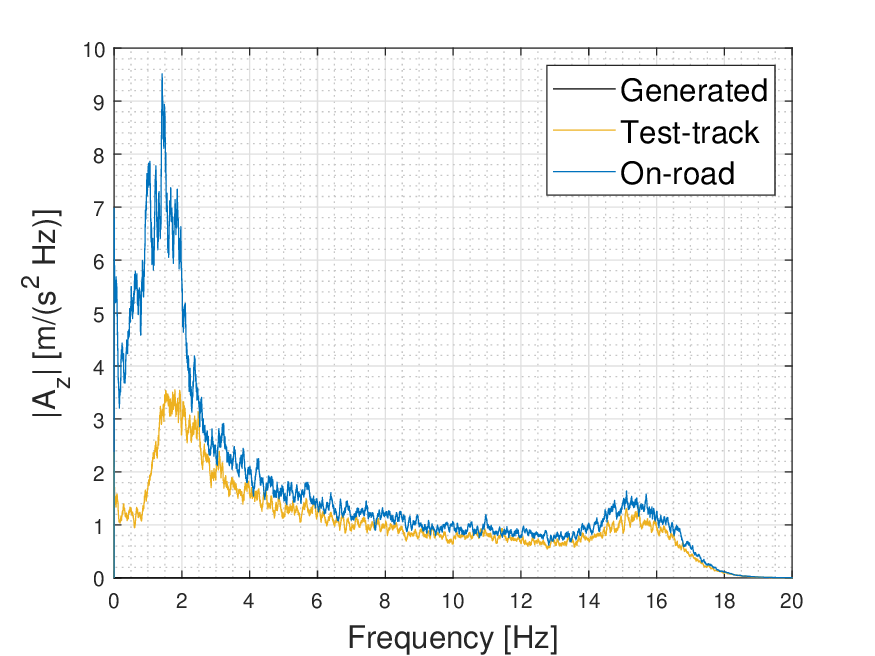}
  \caption{Unweighted $a_z$.}
  \label{fig:az_unfilt}
\end{subfigure}%

\begin{subfigure}{0.33\linewidth}
  \centering
  \includegraphics[width=\linewidth]{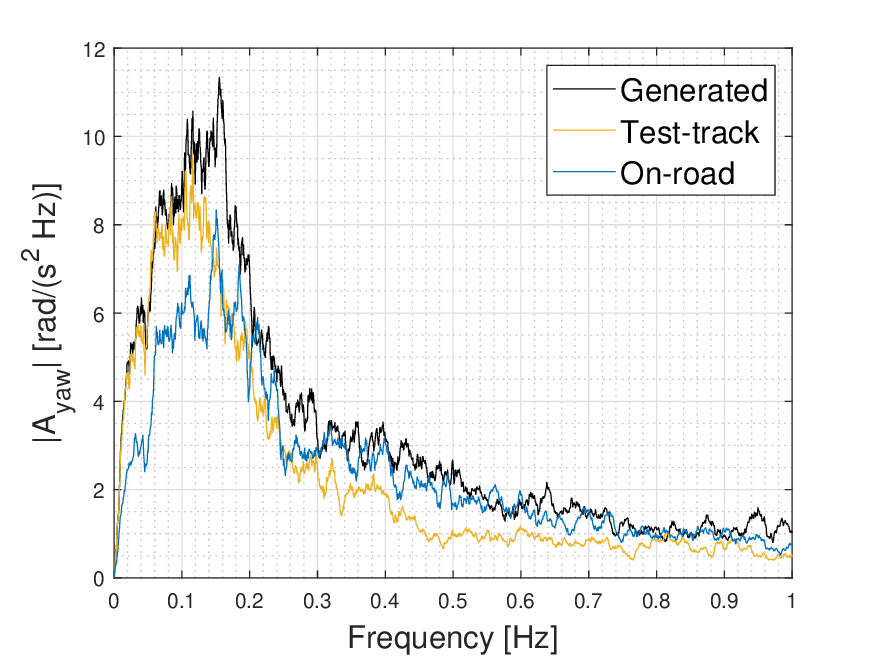}
  \caption{Unweighted $\ddot{r}$.}
  \label{fig:r_unfilt}
\end{subfigure}
\begin{subfigure}{0.33\linewidth}
  \centering
  \includegraphics[width=\linewidth]{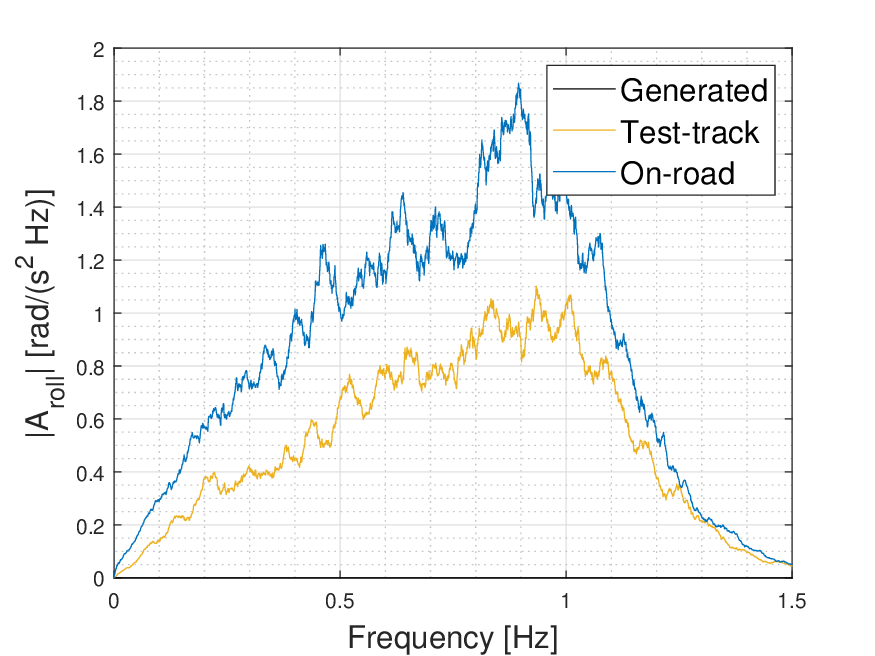}
  \caption{Unweighted $\ddot{\phi}$.}
  \label{fig:roll_unfilt}
\end{subfigure}%
\begin{subfigure}{0.33\linewidth}
  \centering
  \includegraphics[width=\linewidth]{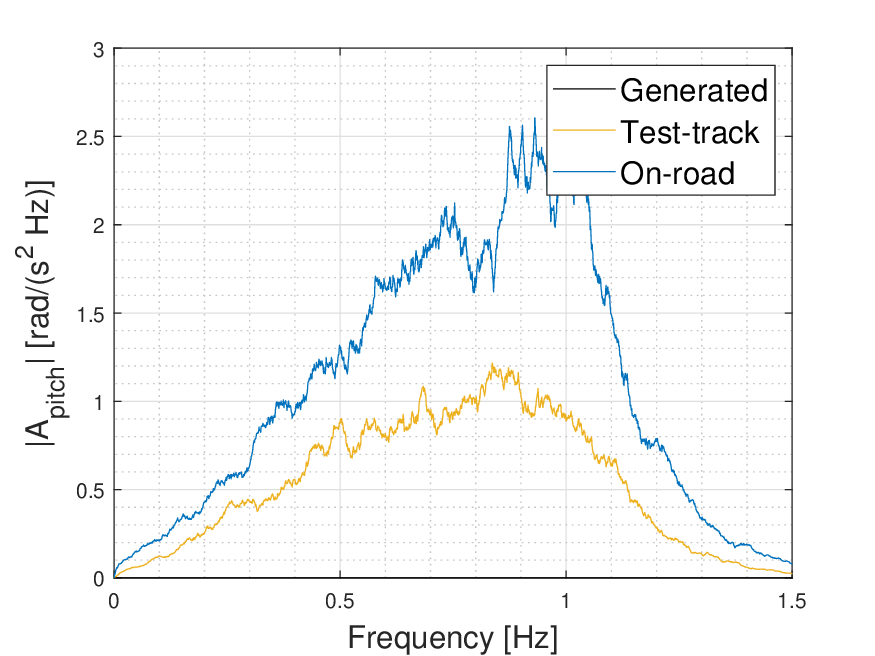}
  \caption{Unweighted $\ddot{\theta}$.}
  \label{fig:pitch_unfilt}
\end{subfigure}
\caption{Single-sided amplitude unweighted accelerations from the generated, on-road and test-track data. The test-track and road data correspond to Participant 33.} %(\ref{fig:ax_unfilt}) $A_{x}$, (\ref{fig:ay_unfilt}) $A_{y}$, (\ref{fig:az_unfilt}) $A_{z}$, (\ref{fig:r_unfilt}) $A_{r}$, (\ref{fig:roll_unfilt}) $A_{roll}$, and (\ref{fig:pitch_unfilt}) $A_{pitch}$.}
\label{fig:freq_plots1}
\end{figure*}

\begin{figure*}[hp!]
\centering
\begin{subfigure}{0.33\linewidth}
  \centering
  \includegraphics[width=\linewidth]{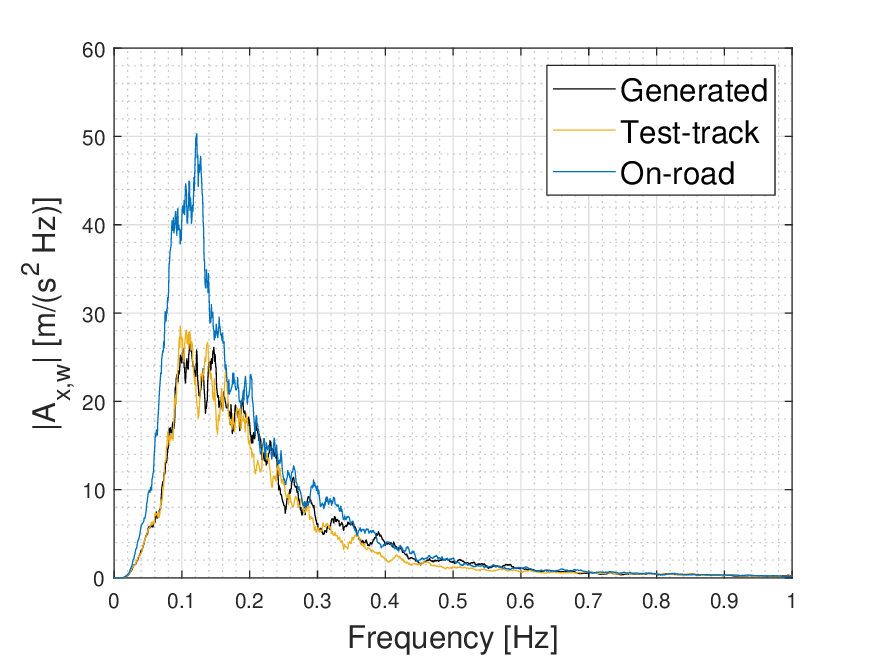}
  \caption{MS-weighted $a_x$.}
  \label{fig:ax_filt}
\end{subfigure}%
\begin{subfigure}{0.33\linewidth}
  \centering
  \includegraphics[width=\linewidth]{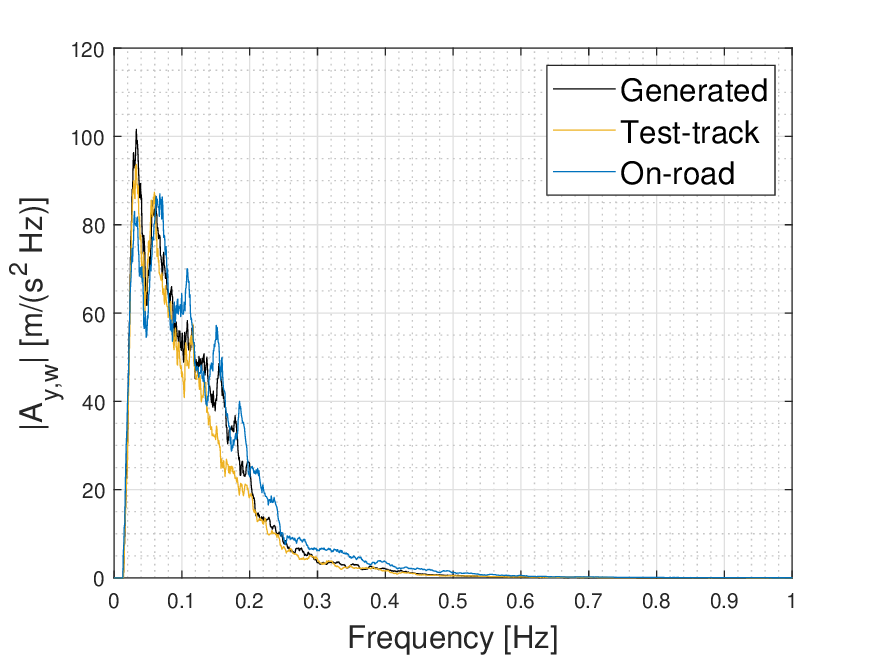}
  \caption{MS-weighted $a_y $.}
  \label{fig:ay_filt}
\end{subfigure}%
\begin{subfigure}{0.33\linewidth}
  \centering
  \includegraphics[width=\linewidth]{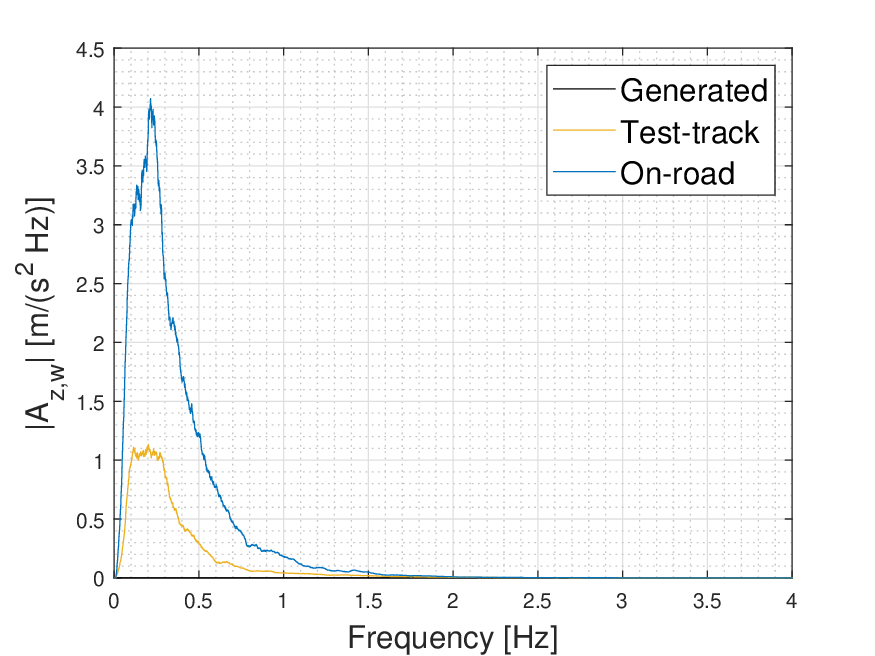}
  \caption{MS-weighted $a_z$.}
  \label{fig:az_filt}
\end{subfigure}%

\hfill
\begin{subfigure}{0.33\linewidth}
  \centering
  \includegraphics[width=\linewidth]{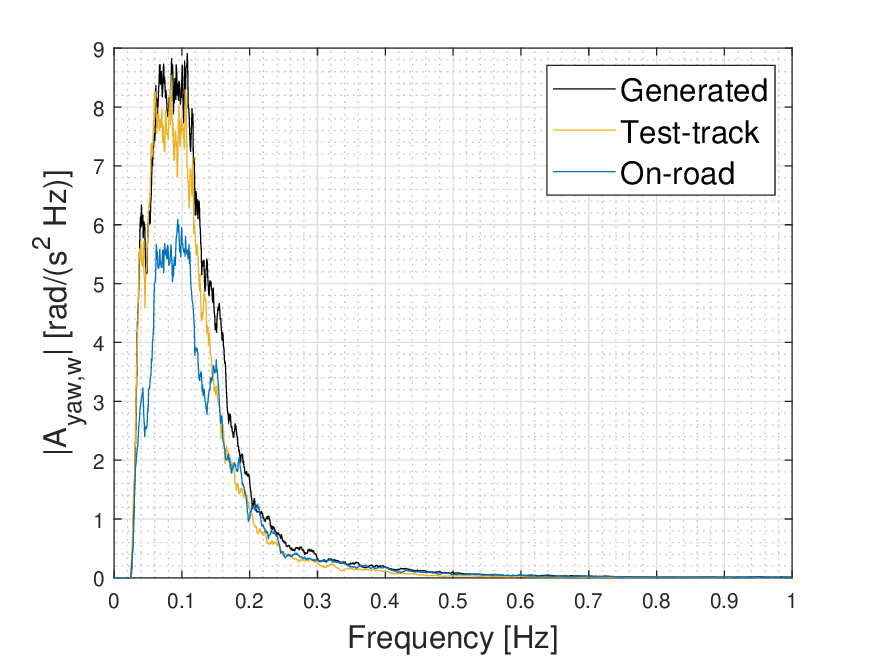}
  \caption{MS-weighted $\ddot{r}$.}
  \label{fig:r_filt}
\end{subfigure}
\hfill
\begin{subfigure}{0.33\linewidth}
  \centering
  \includegraphics[width=\linewidth]{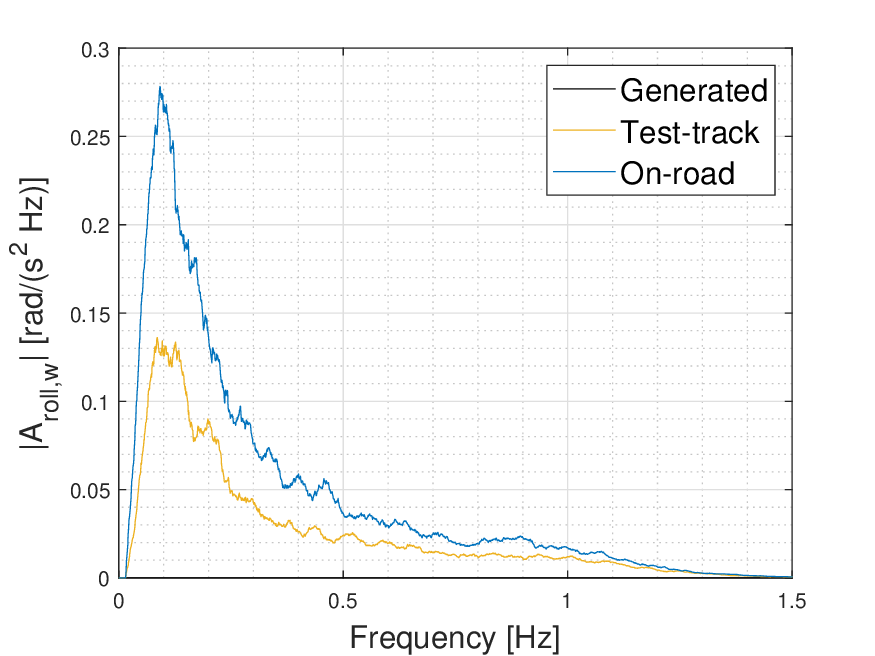}
  \caption{MS-weighted $\ddot{\phi}$.}
  \label{fig:roll_filt}
\end{subfigure}%
\begin{subfigure}{0.33\linewidth}
  \centering
  \includegraphics[width=\linewidth]{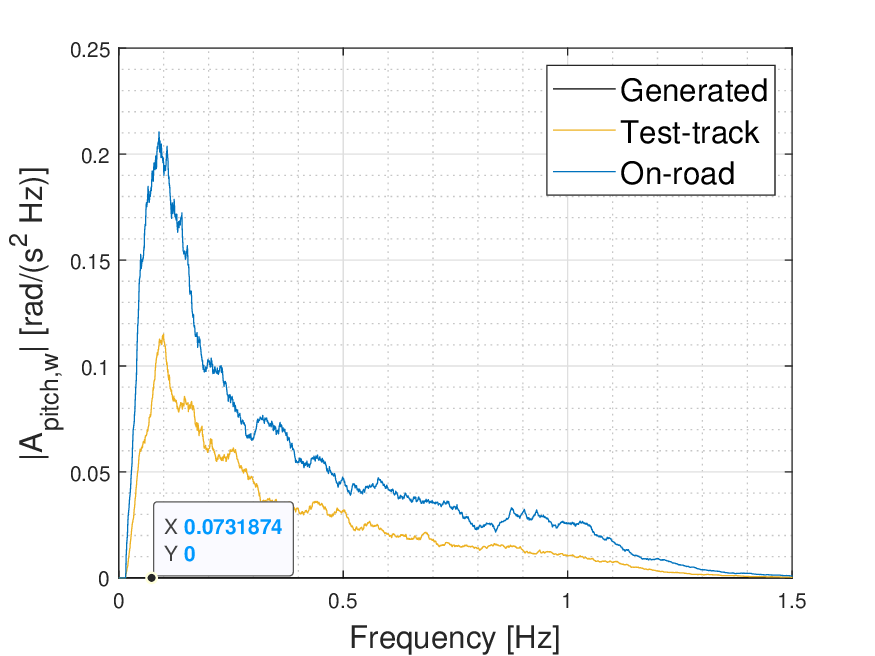}
  \caption{MS-weighted $\ddot{\theta}$.}
  \label{fig:pitch_filt}
\end{subfigure}
\caption{Comparison of single-sided amplitude MS-weighted accelerations from the generated, on-road and test-track data. The test-track and road data correspond to Participant 33.}
\label{fig:freq_plots2}
\end{figure*}

To further assess the tracking performance, the accelerations are compared in the frequency domain.
Figure \ref{fig:ax_unfilt}, \ref{fig:ay_unfilt} and \ref{fig:r_unfilt} present the measured data filtered only with a moving average filter. 
Figure \ref{fig:ax_filt}, \ref{fig:ay_filt} and \ref{fig:r_filt} illustrate the measured data filtered also with regards to motion sickness as detailed in Section \ref{sec:ms_metrics}.

According to Figure \ref{fig:ax_unfilt}, the generated and test-track data are similar both for the longitudinal and lateral accelerations. 
More specifically, the motion planner (generated-data) could not perfectly replicate the low-frequency longitudinal ($|A_{x}|$) of the on-road data due to the compromises made by the algorithm to consider the test track boundaries.
This difference is estimated maximum around $\sim$25$\%$.
The limited space of the test track reduces the controller's ability to fully replicate the on-road longitudinal dynamics. 
Meanwhile, as discussed before, the algorithm prioritises the lateral accelerations. 
Hence, the lateral accelerations ($|A_{y}|$, Figure \ref{fig:ay_filt}) are well replicated.
Furthermore, the implementation of the generated path in the experimental vehicles' trajectory planner, does not illustrate any loss of information. 
However, we identify a notable difference (maximum $\sim$37$\%$) in the yaw accelerations ($|A_{yaw}|$, Figure~ \ref{fig:r_unfilt}), as expected due to the vehicle directional changes to replicate the on-road driving within the boundaries of the test-track.

Regarding the rest of the accelerations, which were not included in the MPC framework, significant differences are identified between on-road and test-track. 
The vertical accelerations ($|A_{z}|$) illustrate significant differences (maximum $\sim$62$\%$) around the low frequencies (0 - 2 Hz), and no loss of information for higher frequencies (2 - 20 Hz).
This is expected due to the different road roughness of the test-track and the on-road. 
The low frequency differences are more provocative for the MS occurrence, and will be explored further regarding their impact.
However, their amplitude is much lower compared to the longitudinal and lateral accelerations. 
As far as the roll ($|A_{\phi}|$) and the pitch ($|A_{\theta}|$) accelerations are concerned, they also illustrate significant differences (maximum $\sim$47$\%$ and $\sim$62$\%$, respectively).
Regarding the roll, the lower amplitude compared to the on-road scenario might be related to the lower speed applied in the test-track. 
Despite the lower amplitude of the roll motion, it is expected to be more frequent in the test-track due to the multiple vehicle directional changes that the vehicle required to replicate the on-road driving scenario. 
Meanwhile, the difference in the pitch accelerations originated from the algorithms inability to fully replicate the longitudinal dynamics. 
Nevertheless, these states together with the vertical accelerations are not considered and controlled with the current MPC framework.
Meanwhile, they are of low amplitude and are not expected to affect the overall motion sickness. 
For this, further analysis is conducted by MS weighted filtering the acceleration as detailed in Section \ref{sec:ms_metrics}.
The identified differences in the raw data of the accelerations, are also illustrated when the MS-weightings are applied (Figure \ref{fig:freq_plots2}). 
The yaw and longitudinal accelerations (Figure \ref{fig:ax_filt}-\ref{fig:r_filt}) still illustrate a significant difference, which is decreased in amplitude due to the filters.

\subsection{Objective comparison of motion sickness assessment}

For further investigation of the differences of these three cases, this section compares the objective motion sickness scores (MSDV metrics, Section \ref{sec:ms_metrics}) based on the vehicle accelerations for the three cases (on-road, generated, and test track data).
More specifically, Table \ref{tab:ms_rc_scores} illustrates the MSDV metric calculated per acceleration signal ($MSDV_i$, with $i=$x, y, z, $\psi$, $\phi$, and $\theta$) and the percentage of difference from the on-road data.
The gray rows in Table \ref{tab:ms_rc_scores} illustrated the signals matched in the MPC framework. 
To generate the track motion, the vertical, roll and pitch dynamics were not considered. Here we also report these signals to quantify their impact on motion sickness. 

According to Table \ref{tab:ms_rc_scores}, similar MSDV metrics are obtained for all three cases illustrating the algorithms ability to effectively replicate road sickness exposure.
The MSDV values are also representative and corresponding to the literature \cite{kato2006study,jain2023motion,metzulat2024does}. 
The on-road data's MSDV score based on the longitudinal accelerations (MS$_x$) shows a difference greater than 33$\%$ compared to the generated and test track data, respectively.
This illustrates how the differences identified in Figures \ref{fig:ax_filt} and \ref{fig:ax_unfilt} translate in motion sickness levels.
Furthermore, the MSDV score based on the lateral accelerations shows smaller differences especially when comparing the generated and the on-road data ($\sim$6$\%$). 
This difference increases a bit due to the change of the acceleration by the experimental vehicle's motion planner, when the generated path was applied on the test-track.
The MSDV score based on the yaw accelerations has increased in the generated data by 38-50$\%$ compared to on-road.
However, the amplitude is much lower compared to the MSDV$_x$ and MSDV$_y$. 
Hence, the impact of the increase in the yaw accelerations is limited on the total MS score (MSDV$_{t}$).
Similarly, the impact of the vertical accelerations is expected to be very small due to the low amplitude despite the significant difference between on-road and test-track ($\sim$72$\%$). 
The MSDV scores based on the roll and pitch accelerations are also significantly higher in the on-road than the test-track but their low amplitude marginally affects the MSDV$_{t}$.

\definecolor{Gray}{gray}{0.85}

\begin{table}[h]
    \centering
    \caption{Objective assessment of motion sickness in the generated, test-track and on-road data through MSDV. The test-track and road data correspond to Participant 33. The difference  of each case with the on-road data is presented as a percentage ($\%$). The ``x" represent non measured signals, since the generated path by the motion planner did not include vertical, roll and pitch motion.}
    \begin{tabular}{|c|c|c|c|c|c|c|} \hline
         % \multicolumn{4}{|c|}{\textbf{Motion sickness assessment}} \\ \hline \hline
         & \textbf{Generated}  & $\%$ & \textbf{Test-track} &  $\%$  & \textbf{On-road}\\ 
         \hline
         \rowcolor{Gray}
         \textbf{MSDV$_x$} [m/s$^{1.5}$]           & 7.28   & 33 & 6.83  & 37  & 10.83 \\ 
         \hline
         \rowcolor{Gray}
         \textbf{MSDV$_y$} [m/s$^{1.5}$]          & 21.94  & 6  & 18.48 & 11  & 20.70 \\ 
         \hline
         \textbf{MSDV$_z$} [m/s$^{1.5}$]          & x      & x  & 0.45  & 71  & 1.52 \\ 
         \hline     
         \rowcolor{Gray}
         \textbf{MSDV$_r$} [rad/s$^{1.5}$]        & 2.01   & 50 & 1.83  & 38  & 1.33 \\ 
         \hline
         \textbf{MSDV$_{\phi}$} [rad/s$^{1.5}$]   & x      & x  & 0.03  & 48  & 0.07 \\ 
         \hline
         \textbf{MSDV$_{\theta}$} [rad/s$^{1.5}$] & x      & x  & 0.04  & 49  & 0.08 \\ 
         \hline        
         \textbf{MSDV$_{t}$}  [m/s$^{1.5}$] & 23.19  & 1  & 19.79 & 17  & 23.44 \\ 
         \hline 
    \end{tabular}
    \label{tab:ms_rc_scores}
\end{table}

\section{Results: Part B - Motion sickness assessment}
\label{sec:resultsB}

\subsection{Objective assessment using experimental data}

\begin{figure}[h]
    \centering    
    \includegraphics[width=1\linewidth]{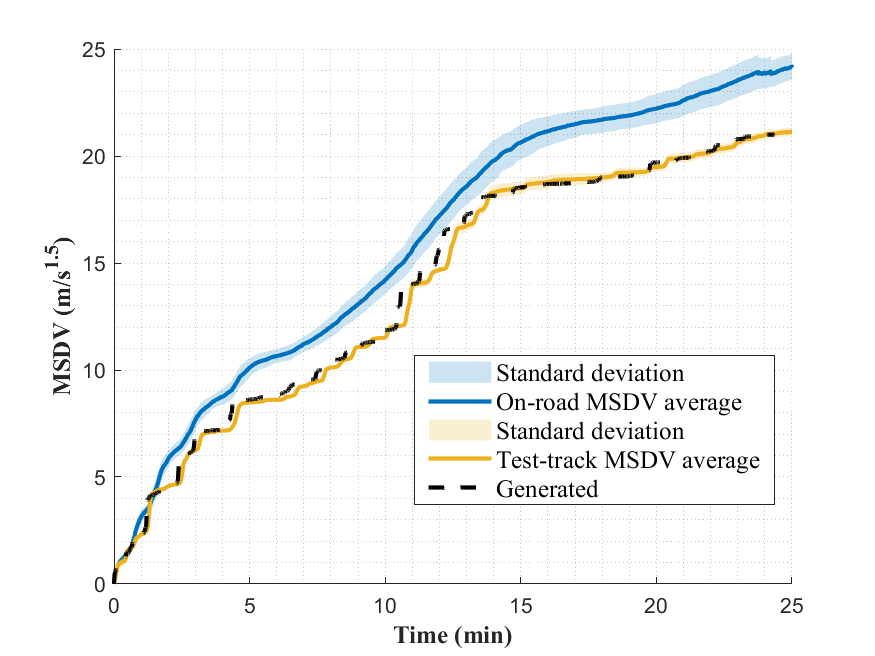}
    \caption{Motion sickness exposure (MSDV) over 47 participant runs on-road and test-track, and for the generated path.}
    \label{fig:MSDV_total}
\end{figure}

According to Table \ref{tab:ms_rc_scores}, the contribution of MS$_z$, MS$_\psi$, MS$_\phi$ and MS$_\theta$ to the MS$_{total}$ were small (i.e., less than 10$\%$ in total) both for the test-track and on-road data. 
This also aligns with the literature, where similar results were identified regarding the limited impact of the rotational accelerations on motion sickness objective metrics \cite{papaioannou2023impact, Howarth2003}. 
Therefore, during the complete human participant experiments, only longitudinal and lateral acceleration were recorded and presented. 
Hence, the MSDV scores were calculated by considering only these two accelerations. 
The mean MSDV scores representing the sickness exposure of the applied motion are shown in Figure \ref{fig:MSDV_total}. 
The blue line depicts the mean MSDV for the on-road condition, whereas the yellow green line shows the mean MSDV for the test-track condition. 
The shaded areas around the plotted lines depict the standard deviation of MSDV.
Meanwhile, the MSDV is also calculated and plotted for the generated and the reference path.  
Based on the results, the averaged MSDV for all the participants in the test track condition is perfectly aligned with the generated path from the motion planner. 
This illustrated that the vehicle's path tracking was able to track the generated path by our motion planner.
However, according to Figure \ref{fig:MSDV_total}, there is an average $\sim$12$\%$ difference between test-track and on-road in the averaged MSDV values of all participants. 
This difference is also significant (p $<$ .001) according to a Mann-Whitney U Test. 
Therefore, based on the MS objective metrics we identify a small ($\sim$12$\%$) decrease in the test-track compared to on-road.

\subsection{Subjective assessment by human participants}

Having objectively explored the efficiency of the algorithm, we subjectively assessed if the change of participants' MISC levels over time in the two conditions differs.
This is tested using the participants' subjective input regarding motion sickness using MISC scores. 
During both conditions, the participants were engaged in watching sports videos as NDRT. 
According to the NDRT\_counts, their focus during the ride was similar in both conditions and relatively high.
More specifically, the participants identified on average 15.81 and 14.70 NDRT\_counts out of 16 in total during the on-road and test-track condition.

The participants' high focus on the NDRT in both conditions secured that they had their eyes-off the road. 
Therefore, external vision and the external environment will not have affected the accumulation of MS and the MISC levels.
The mean value of the maximum MISC levels was 2.69 and 2.29 for the road and test-track condition, respectively. 
This 15$\%$ reduction difference is close to the  objective MSDV metric (Figure \ref{fig:MSDV_total}), where the test-track had a 12$\%$ lower sickness exposure compared to on-road. 
The maximum MISC levels between the two conditions did not illustrate any significant difference (p~=~.53) according to a Mann-Whitney U test. 

\begin{figure}[h]
    \centering
    \includegraphics[width=1\linewidth]{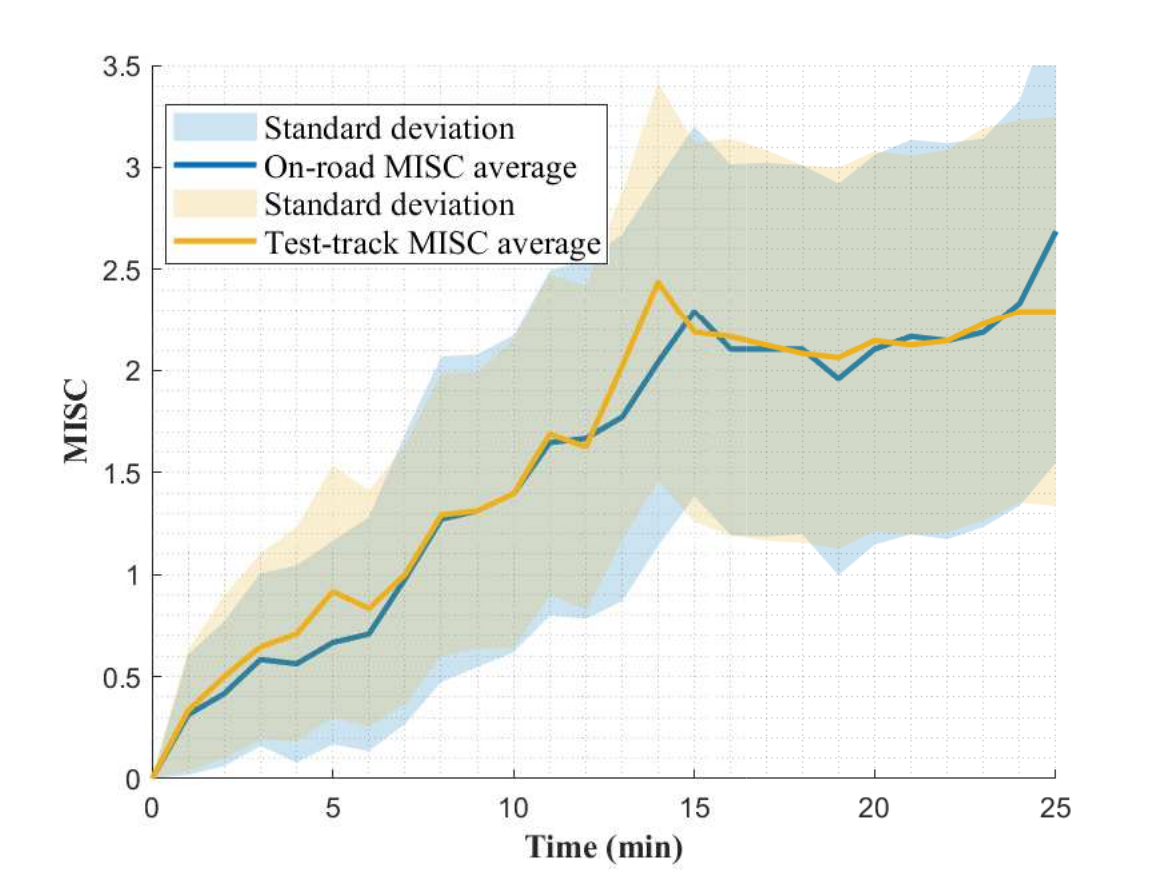}
    \caption{Subjective motion sickness assessment (MISC) over 47 participants for on-road and test-track.}
    \label{fig:avg_misc}
\end{figure}

The MISC scores collected from the participants over time while being driven are presented in Figure \ref{fig:avg_misc}. 
The values were asked at the beginning of the drive and then every minute at each point in time (x-axis). 
A termination criterion for our experiment during both conditions was set if participants reached a MISC score of 6 or higher. 
The repeated measures ANOVA with time and condition as a factor showed a significant effect of time on the MISC response (F(2,48)=70.0, p $<$ .001), proving that carsickness accumulated over time. 
Meanwhile, the ANOVA on MISC as function of time also shows no effect of the condition (F(2,48) = 0.79, p = .75) on the occurrence of MISC. 
Thereby the MISC on track does not differ significantly from on-road.

\begin{figure}[!h]
    \centering    \includegraphics[width=1\linewidth]{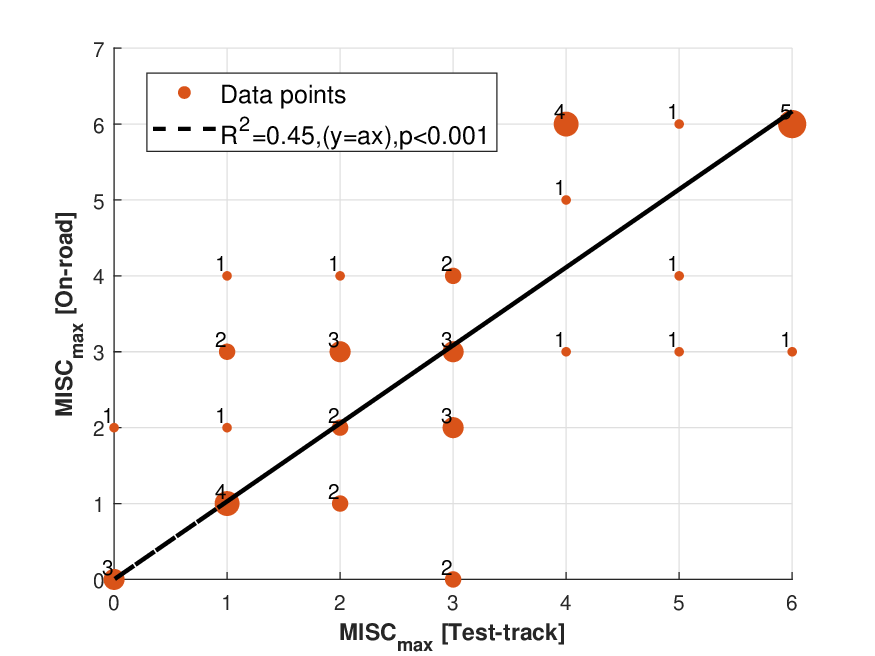}
    \caption{Correlation of individual MISC$_{max}$ between the two conditions. The numbers of duplicates per co-ordinate are illustrated in the figure, while the size of the points changes accordingly. The data are fitted in first order polynomial functions (y = a * x).}
    \label{fig:MISC_total_correlation}
\end{figure}

In addition to the effective replication of the average motion sickness occurrence, the replication of the individual motion sickness occurrence is also critical. 
Hence, we opt to explore this through Figure \ref{fig:MISC_total_correlation}, where we see a good correspondence at the extremes and at low levels of MISC.
More specifically, three participants reported no sickness both on road and on track and five other participants reached the termination criterion (MISC=6) both on road and on track.
Also, four and two participants reported level 1 and 2 MISC in both conditions, respectively . 
The figure presents the correlation of the individual MISC$_{max}$ between two conditions, fitting the data to a first order polynomial (y=a*x). 
The optimal fit is achieved with a = 1.03, with an adjusted $R^2$ = 0.45 (p$<$0.001).

\section{Limitation and Future Work}

Despite the proven efficiency of the method, there are certain limitations:
\begin{itemize}
    \item \textbf{Test-track space limitations:} The reproducability of the longitudinal accelerations in the test-track would have been higher if more space was available. 
    However, even if this difference was captured by objective metrics, there was no significant difference between the participant's responses during the test-track and the on-road scenarios.
    To further enhance the replication of on-road exposure more advanced MPC approaches can be used, such as the frequency-splitting MPC \cite{jain2023motion}.
    
    \item \textbf{Testing environment:} Although we showed that the algorithm can replicate the motion sickness exposure, the effect of the two different testing environments as well as personality traits on the individual responses cannot be ruled out.
    Therefore, further work is in progress to exploit the data collected from this experiment to unravel this aspect. 
    
    \item \textbf{Motion sickness occurrence:}  Our method was proven effective in replicating on-road driving scenarios with low MISC levels, but it is expected to be able to also replicate on-road driving scenarios with higher MISC levels.
    
    \item \textbf{Engagement in NDRT:} The participants were instructed to focus on the non-driving related task of video watching. 
    This ensured participants’ exposure to an environment similar to using an AV in an eyes-off-road mode. In practice, AV users will occasionally observe the road, and non-driving tasks may vary in visual and motor complexity, which have a modest but significant effect on MS \cite{metzulat2024does}.    
    Hence, further work by the authors is in progress to prove the generalization of the method for efficiently assessing motion sickness while not engaged in NDRTs and other conditions.
    \item \textbf{Equipment requirements:}  Even if this method allows the effective replicability of any driving scenario, an automated vehicle or steering and pedal robots are needed for the vehicle to follow the generated path.    
    
\end{itemize}

%%%%%%%%%%%%%%%%%%%%%%%%%%%%%%%%%%%%%%%%%%

\section{Conclusions}

\label{sec:conclusion}

In this work, a motion planner using model predictive control has been developed to replicate on-road sickness stimuli on a small test track.
The algorithm outputs a reference velocity and trajectory profile for the vehicle to follow within the test-track boundaries.
The effectiveness of the algorithm is proven by comparing objective metrics of motion sickness stimuli in time in terms of motion sickness dose value and frequency domain between the on-road, generated, and test track data.  
Furthermore, a within-subjects experiment with 47 human participants was conducted to test if the motion sickness occurrence did not differ between the two conditions.
The results illustrated that our motion planner is able to replicate an on-road driving scenario in a small test track by maintaining highly similar motion sickness stimuli with lower speeds. 
This breakthrough proves that data collection can become more efficient and within a safer and controllable environment, allowing to speed up research into fundamentally understanding motion sickness in automated vehicles, modelling and prediction of motion sickness, and developing its mitigation strategies.
This method enables testing of early stage prototype or unreleased developed stages without approval of driving on pubic roads, securing realistic results with regards to on-road driving.
Further work is in progress to exploit the collected data to unravel other differences that might rise between test-track and on-road testing.

\section*{Acknowledgments}

The research leading to these results has received funding from the European Union Horizon 2020 Research and Innovation Programme under grant agreement No 101006664 (Hi-Drive). The project received additional financing by Volkswagen AG. 
We would also like to thank our Hi-Drive partners for their support and feedback. 
Special thanks to Mr Aaron Edelmann, Ms Myriam Metzulat, Ms Hanna Körber for their support. 
They all contributed in the design of the experiment for the future steps of this work, and Ms Hanna Körber additionally supported in the data collection actively. Additionally, we thank Marcel Niklas for the technical support in the realisation of the trajectory driving function. 

\ifCLASSOPTIONcaptionsoff
  \newpage
\fi

\bibliography{00_Main}
%\bibliography{Project}
\bibliographystyle{IEEEtran}

\begin{IEEEbiography}[{\includegraphics[width=1in,height=1.25in,clip,keepaspectratio]{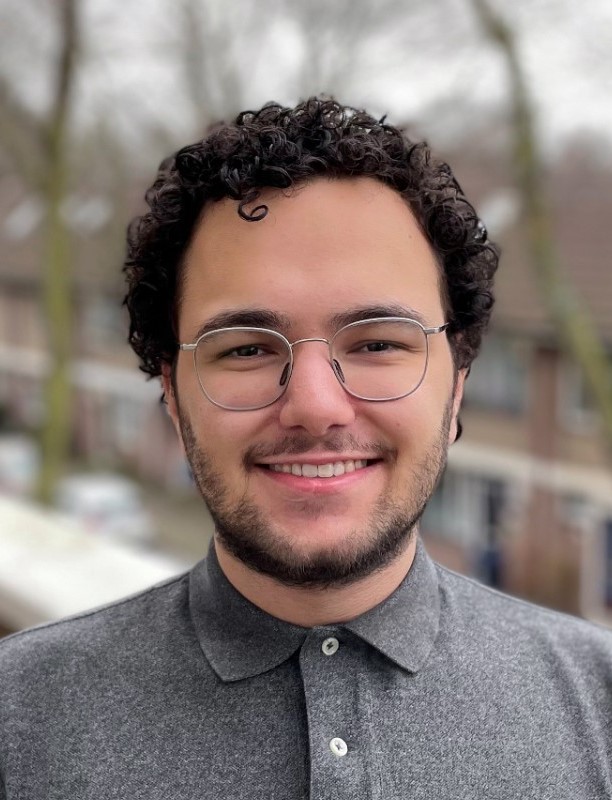}}]%
{Hüseyin Harmankaya} received a MSc degree, 2023, in Robotics from Delft University of Technology. He joined the Intelligent Vehicles group, Delft University of Technology, in 2023 as a Researcher focusing on motion comfort and vehicle dynamics and control.
\end{IEEEbiography}

\begin{IEEEbiography}[{\includegraphics[width=1in,height=1.25in,clip,keepaspectratio]{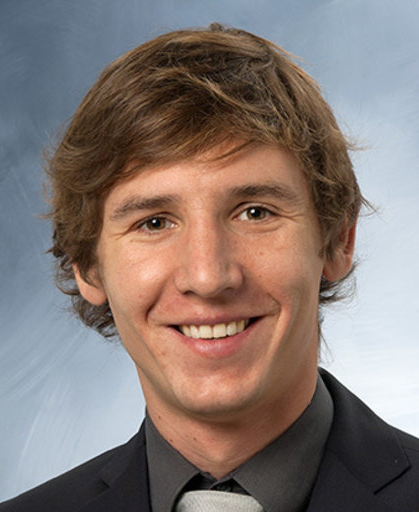}}]%
{Adrian Brietzke} received a PhD degree, 2022, in Engineering from Chemnitz University of Technlogy in cooperation with Volkswagen AG. He is a Researcher in the field of Motion Comfort since 2011 in the Volkswagen Group Innovation. His focus lies on the analysis of vehicle and human body dynamics as well as the development of automotive prototypes.  
\end{IEEEbiography}

\begin{IEEEbiography}[{\includegraphics[width=1in,height=1.25in,clip,keepaspectratio]{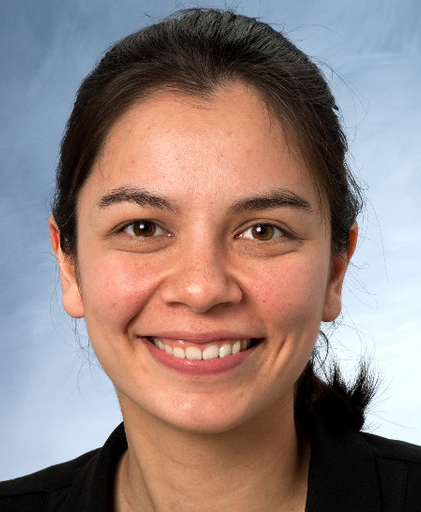}}]%
{Rebecca Pham Xuan} received a PhD degree, 2022, in Engineering from  Technical University Berlin in cooperation with Volkswagen Group Innovation. She researches and promotes innovations in the field of motion sickness at the Volkswagen Group Innovation in Wolfsburg.
Her interests lie in the analysis of data collected in a real driving environment and the user-centered development of mitigation methods.
\end{IEEEbiography}

\begin{IEEEbiography}[{\includegraphics[width=1in,height=1.25in,clip,keepaspectratio]{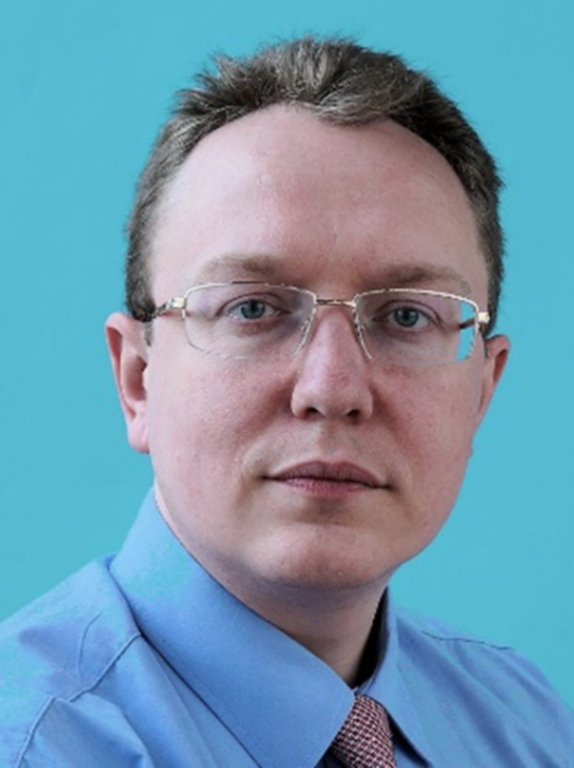}}]%
{Barys Shyrokau} received a joint PhD degree, 2015, in Control Engineering from Nanyang Technological University and Technical University Munich. He is an associate professor in the section of Intelligent Vehicles at Delft University of Technology, and his research interests are vehicle dynamics and control, motion comfort, and driving simulator technology. Scholarship and award holder of DAAD, IFAC, FISITA, SAE, SINGA, ISTVS, and CADLM.
\end{IEEEbiography}

\begin{IEEEbiography}[{\includegraphics[width=1.1in,height=1.25in,clip,keepaspectratio]{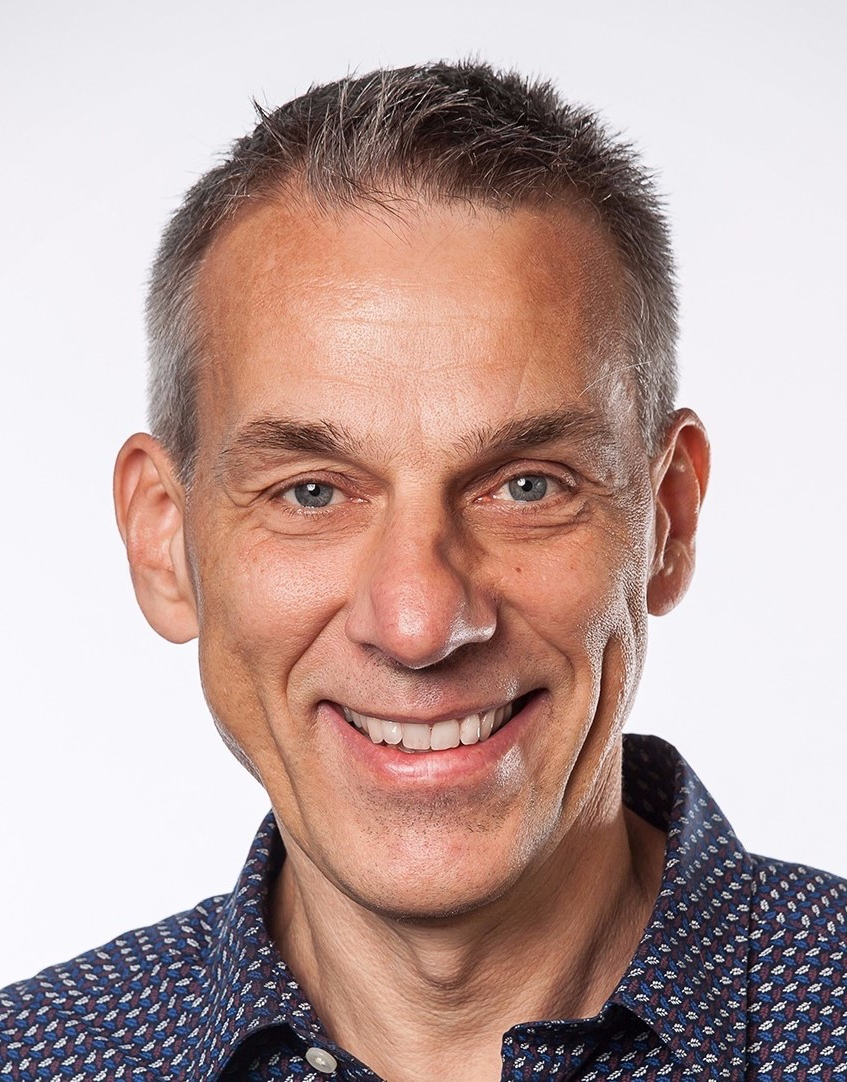}}]%%
{Riender Happee} received the Ph.D. degree from TU Delft, The Netherlands, in 1992.
He investigated road safety and introduced biomechanical human models for impact and comfort at TNO Automotive (1992-2007). Currently, he investigates the human interaction with automated vehicles focusing on motion comfort, perceived safety and acceptance at the Delft University of Technology, the Netherlands, where he is full Professor.
\end{IEEEbiography}

\begin{IEEEbiography}[{\includegraphics[width=1.1in,height=1.25in,clip,keepaspectratio]{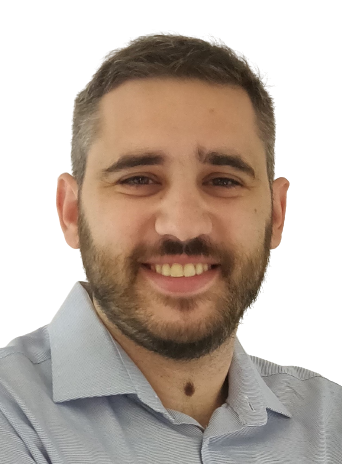}}]%
{Georgios Papaioannou} received the Ph.D. degree from the National Technical University of Athens (NTUA), Greece, in 2019, which received an award regarding its innovation and impact. He is currently an Assistant Professor on motion comfort in AVs at TU Delft, after conducting postdoctoral research at KTH Royal Institute of Technology in Sweden and Cranfield University in U.K. His research interests include human factors, motion comfort, seat comfort, postural stability, human body modelling, automated vehicles, remote driving, dynamics and control, and extended reality.
\end{IEEEbiography}

\end{document}